\documentclass[journal]{IEEEtran}

\ifCLASSINFOpdf
\else
\usepackage[dvips]{graphicx}
\fi
\usepackage{url}
\usepackage{float} 
\usepackage{graphicx}
\usepackage{amsmath}
\usepackage{amssymb}
\usepackage{booktabs}
\usepackage{threeparttable}
\usepackage{multirow}
\usepackage{color}
\usepackage[OT1]{fontenc}
\usepackage{cite}
\usepackage{algorithmic}
\usepackage{algorithm}
\usepackage{multirow}

\usepackage{amsthm}

\usepackage{setspace}
\usepackage{hyperref}
\hypersetup{hidelinks,
	colorlinks=true,
	allcolors=blue,
	pdfstartview=Fit,
	breaklinks=true}

\usepackage{easyReview}

\begin{document}
	
	\title{Adaptive Modality Balanced Online Knowledge Distillation for Brain-Eye-Computer based \\Dim Object Detection}
	\author{Zixing Li, Chao Yan, Zhen Lan, Xiaojia Xiang, Han Zhou, Jun Lai, Dengqing Tang
		
	\thanks{Zixing Li, Zhen Lan, Xiaojia Xiang, Han Zhou, Jun Lai, Dengqing Tang are affiliated with the College of Intelligence Science and Technology, National University of Defense Technology, Changsha 410073, China, (e-mail: \{lizixing16, lanzhen19, xiangxiaojia, zhouhan, laijun, tangdengqing09\} @nudt.edu.cn). Zixing Li and Chao Yan contributed equally to this work.\textit{(Corresponding author: Dengqing Tang.)}}

	}
	\markboth{IEEE Transactions on Neural Networks and Learning Systems}
	{Shell \MakeLowercase{\textit{\emph{et al.}}}: Bare Demo of IEEEtran.cls for IEEE Journals}
	\maketitle
	
	\begin{abstract}
		Advanced cognition can be measured from the human brain using brain-computer interfaces. Integrating these interfaces with computer vision techniques, which possess efficient feature extraction capabilities, can achieve more robust and accurate detection of dim targets in aerial images. However, existing target detection methods primarily concentrate on homogeneous data, lacking efficient and versatile processing capabilities for heterogeneous multimodal data. In this paper, we first build a brain-eye-computer based object detection system for aerial images under few-shot conditions. This system detects suspicious targets using region proposal networks, evokes the event-related potential (ERP) signal in electroencephalogram (EEG) through the eye-tracking-based slow serial visual presentation (ESSVP) paradigm, and constructs the EEG-image data pairs with eye movement data. Then, an adaptive modality balanced online knowledge distillation (AMBOKD) method is proposed to recognize dim objects with the EEG-image data. AMBOKD fuses EEG and image features using a multi-head attention module, establishing a new modality with comprehensive features. To enhance the performance and robust capability of the fusion modality, simultaneous training and mutual learning between modalities are enabled by end-to-end online knowledge distillation. During the learning process, an adaptive modality balancing module is proposed to ensure multimodal equilibrium by dynamically adjusting the weights of the importance and the training gradients across various modalities. The effectiveness and superiority of our method are demonstrated by comparing it with existing state-of-the-art methods. Additionally, experiments conducted on public datasets and real-world scenarios demonstrate the reliability and practicality of the proposed system and the designed method. The dataset and the source code can be found at: \href{https://github.com/lizixing23/AMBOKD}{https://github.com/lizixing23/AMBOKD}.
		
	\end{abstract}
	
	\begin{IEEEkeywords}
		object detection, EEG, multimodal learning, online knowledge distillation, adaptive modality balancing
	\end{IEEEkeywords}

	\section*{ABBREVIATIONS}
	\addcontentsline{toc}{section}{Abbreviations}  
	\label{sec:nomenclature}
	
	\begin{table}[!h]
		\label{tab:abbreviations}
		\centering
		\small
		\begin{tabular}{lp{6cm}}
			\toprule
			\textbf{Abbreviation} & \textbf{Full Form} \\
			\midrule
			AMBOKD & Adaptive modality balanced online knowledge distillation \\
			AMB & Adaptive modality balancing \\
			ANOVA & Analysis of variance\\
			BCIs & Brain-computer interfaces \\
			CE & Cross-entropy \\
			EEG & Electroencephalogram \\
			ERP & Event-related potential \\
			ESSVP & Eye-tracking-based slow visual presentation\\
			KD & Knowledge distillation \\
			OKD & Online knowledge distillation \\
			RPN & Region proposal network \\
			RSVP & Rapid serial visual presentation \\
			\bottomrule
		\end{tabular}
		\vspace{-1em}
	\end{table}
	\section{Introduction}
	\IEEEPARstart{I}{n} recent years, considerable progress has been made in the field of computer vision, primarily due to rapid advancements in deep learning~\cite{9451544}. Remarkable results in visual tasks, such as object recognition, image generation, and visual localization, have been achieved through deep learning, leveraging high-performance computing, well-designed neural networks, and large datasets~\cite{10.1109/TPAMI.2023.3290594}. However, accurately and robustly detecting dim objects in aerial images remains challenging due to cluttered backgrounds, varying observing angles, and small object scales~\cite{jia2022low}. Moreover, in target-unknown detection tasks, such as battlefield reconnaissance and disaster rescue, limited training samples from UAVs hinder accurate target detection by purely computer vision-based methods. Human judgment is often required to address these challenges.
	
	While manual methods are inefficient, brain-computer interfaces (BCIs) present a promising solution to these challenges. This technique is able to decode brain activity in response to events, leveraging the prior knowledge and advanced cognitive abilities of humankind. Existing studies have demonstrated the potential of using event-related potential (ERP) detection for target recognition tasks~\cite {lan2021macro,li2022mcgram}. By detecting and interpreting ERPs in electroencephalogram (EEG) signals, these works have shown effectiveness in target recognition tasks across different scenarios~\cite{Lees2018ARO}. However, due to the high noise of the EEG signals, the accuracy of EEG-based target recognition approaches still needs to be enhanced.
	
	Research demonstrates that computer vision systems effectively transform visual stimuli into discriminative semantic representations by leveraging large-scale imaging transformations and modeling intra-class variations~\cite{quan2024multimodal}. In contrast, human visual processing achieves subconscious high-level feature extraction through the dynamic interplay between bottom-up stimulus-driven mechanisms and top-down attentional modulation~\cite{wang2021native}. Multimodal learning approaches integrate these feature representations, leveraging the complementary advantages of computer vision and EEG modalities~\cite{baltruvsaitis2018multimodal}. However, current multimodal fusion approaches~\cite{10024142,nagrani2021attention,yang2022multimodal} typically rely on static weighting strategies, which assign fixed importance coefficients to different modalities throughout the training process. This fundamental assumption contradicts the empirical observation that modality reliability evolves dynamically, particularly evident in EEG-visual fusion where visual features typically converge faster than neural signals. Furthermore, existing knowledge distillation frameworks~\cite{Hinton2015DistillingTK,Zhang2022VisualtoEEGCK,zhang2022confidence} often rely on unidirectional knowledge transfer patterns, neglecting the potential synergistic effects that could emerge from bidirectional cross-modal interactions. These inherent limitations become particularly detrimental in few-shot aerial detection scenarios, where the scarcity of training data amplifies the impact of suboptimal fusion strategies.
	
	\begin{figure}
		\centerline{\includegraphics[width=0.9\columnwidth]{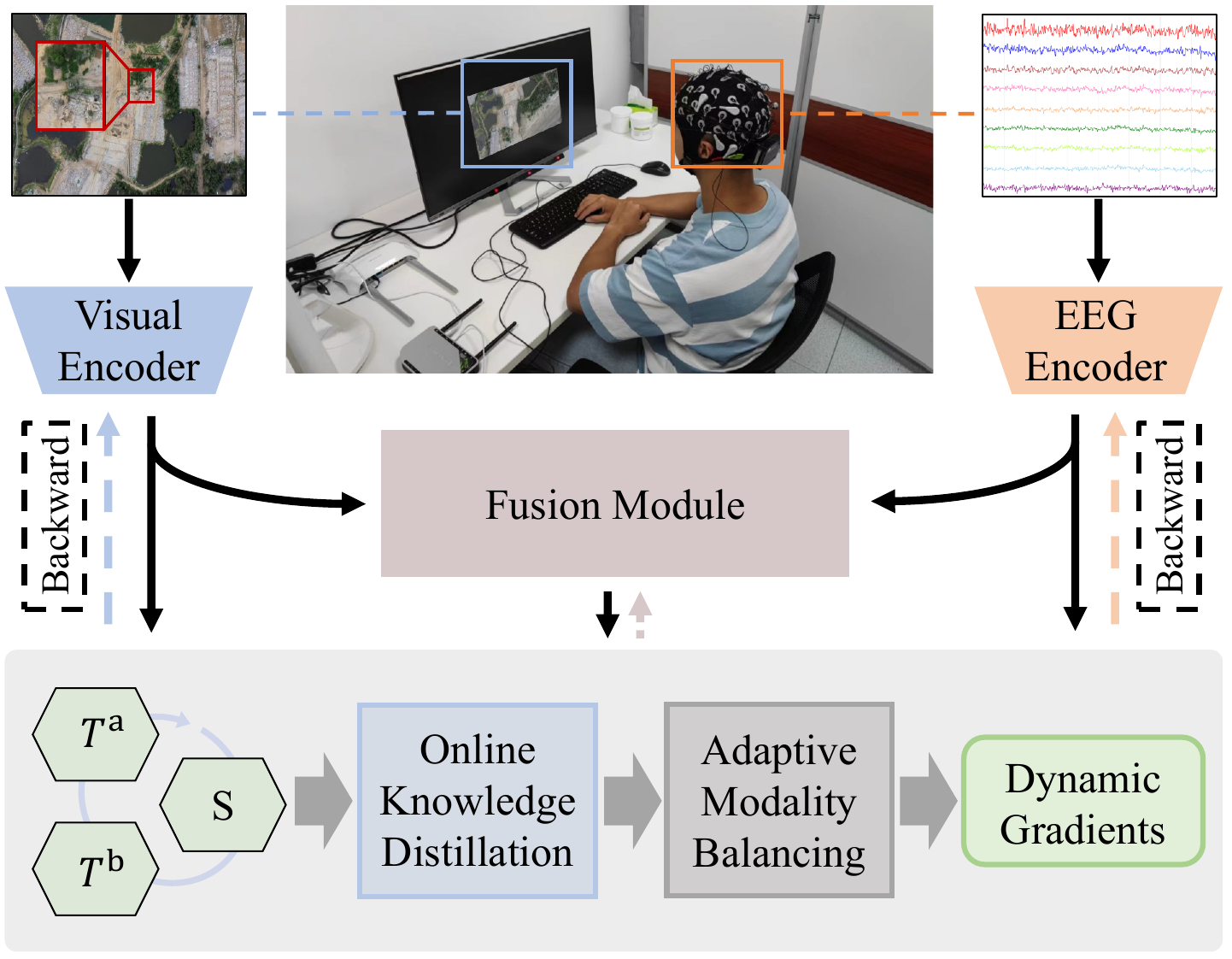}}
		\caption{Brain-eye-computer based object detection system with our proposed AMBOKD method.}
		\label{figurelabel0}
		\vspace{-1em}
	\end{figure}
	
	In this study, we develop a brain-eye-computer-based object detection system for dim objects in aerial images under few-shot conditions, leveraging human cognitive strengths and computer vision’s data processing capabilities. This system employs region proposal networks to detect suspicious object regions, elicits ERP signals using the eye-tracking-based slow serial visual presentation (ESSVP) paradigm, and constructs EEG-image data pairs with eye movement data for target recognition. As shown in Fig.~\ref{figurelabel0}, the proposed adaptive modality balanced online knowledge distillation (AMBOKD) method establishes a tri-modal framework where the fusion model operates as an independent modality, enabling mutual knowledge exchange with both visual and EEG streams. This framework dynamically adjusts modality-specific learning rates and cross-modal knowledge weights in real-time, ensuring balanced optimization across heterogeneous data representations.
	
	To the best of our knowledge, this study is the pioneering effort in addressing the challenges of dim object detection in aerial images using heterogeneous multimodal data, which holds great promise in fewshot scenarios such as military reconnaissance and disaster search and rescue. The main contributions of this study can be summarized as follows:
	
	\begin{itemize}
		\setlength{\itemsep}{0pt}
		\item A novel brain-eye-computer based object detection system is established, by leveraging the complementary strengths of machine vision and human vision, achieving superior performance in detecting dim targets in aerial images.
		\item An AMBOKD method is proposed to fuse heterogeneous multi-modal data, which significantly enhancing detection performance through online mutual learning with dynamic modality balancing.
		\item The first publicly available ESSVP dataset is introduced, which incorporates temporally aligned EEG-visual pairs with aerial imaging characteristics, addressing a crucial gap in multimodal brain-computer vision research.
	\end{itemize}
	
	The rest of this paper is structured as follows: Section~\ref{seq2} discusses the related work of this study. Section~\ref{seq3} presents the design of the built system and the ESSVP dataset in detail. Section~\ref{seq4} illustrates the proposed target recognition method. Section~\ref{seq5} presents the experimental results and analyzes the performance. Section~\ref{seq6} draws conclusions and possible directions for future work.
	\section{Related Work}
	\label{seq2}
	\subsection{Brain-Computer Based Object Recognition}
	
	Computer-based target recognition approaches have been fully developed because of the rapid development of deep learning. The well-known neural network algorithms commonly employed in this domain include ResNet~\cite{he2016deep}, MobileNetV2~\cite{sandler2018mobilenetv2}, and EfficientNet~\cite{tan2019efficientnet}. However, the existing algorithms still suffer from false alarms and missed detection problems. These algorithms are highly dependent on the data set due to the influence of factors such as the target environment, training data and noise. 
	
	In recent years, EEG-based object recognition approaches have attracted considerable attention because of their ability to extract human cognition~\cite{zheng2021attention}. These approaches depend on the analysis of ERP signals generated when subjects identify a target, providing enhanced robustness in intricate environments and minimizing the need for large amounts of image data~\cite{lan2021macro,fan2022dc}. The rapid serial visual presentation (RSVP) paradigm evokes ERP signals employed for target recognition by presenting visual image sequences as stimuli. For instance, Lan \emph{et al.}~\cite{lan2021macro} proposed a multi-attention convolutional recurrent model for ERP detection, facilitating the identification of target images within RSVP sequences~\cite{zhang2020benchmark}. In addition, Fan \emph{et al.}~\cite{fan2022dc} introduced an asynchronous visual evoked paradigm (AVEP) based on RSVP, and proposed a deep learning method for detecting dim objects in satellite images. However, EEG signals are different on individual subjects and are prone to noise interference during signal extraction, thereby necessitating further improvements.
	
	To harness the benefits of EEG and computer vision approaches, the fusion of EEG and image features has been investigated to accomplish efficient and robust object recognition. Barngrover \emph{et al.}~\cite{barngrover2015brain} developed a specific brain-computer Interface (BCI) system that addresses the challenge of target mine recognition inside-scan sonar images by fusing image and EEG signal features. Song \emph{et al.}~\cite{song2024decoding} employed a self-supervised framework to demonstrate the feasibility of learning image representations from EEG signals. Their extensive experiments affirm the biological plausibility of this approach, providing insights into human object recognition from temporal, spatial, spectral, and semantic perspectives. Quan \emph{et al.}~\cite{quan2024multimodal} utilized the rhesus monkey dataset~\cite{schrimpf2020integrative} to analyze the similarities and differences between image representations derived from brain responses and those computed by deep convolutional neural networks, demonstrating the feasibility of EEG-visual fusion. Furthermore, they developed an EEG-visual fusion model to jointly learn brain responses and image representations from convolutional neural networks, validating the model’s performance on a self-collected dataset.
	
	However, existing approaches mainly target salient objects or simple scenarios, neglecting to develop robust cross-modal systems for aerial object detection. To address this gap, our study proposes a brain-eye-computer based system for dim target detection in aerial images. This system detects the suspicious object region with the region proposal networks and employs eye-tracking technology to simultaneously extract the attention regions and EEG signal fragments of the subject from the images. These complementary signals are fused via an adaptive fusion framework, achieving enhanced detection precision, particularly for challenging dim targets.
	
	\subsection{Multimodal Learning}
	Modeling and analyzing data from various sensory modalities, including images, speech, text, and EEG signals, are key aspects of multimodal learning~\cite{10089190,zhou2022mtanet}. Significant attention has been garnered by multimodal learning in the fields of artificial intelligence and machine learning in recent years, resulting in remarkable progress~\cite{baltruvsaitis2018multimodal,10123038}. 
	
	Due to the development of deep learning, researchers propose novel loss functions or training approaches to optimize multimodal models. For instance, Du \emph{et al.}~\cite{10089190} modeled the relationships between brain, visual, and linguistic features using multimodal deep generative models. They maximized both intra-modality and inter-modality mutual information regularization terms. Their approach addresses limitations such as the under-exploitation of multimodal knowledge and the scarcity of training data. Yang \emph{et al.}~\cite{yang2022multimodal} proposed a multimodal fusion approach for remote sensing image-audio retrieval tasks. They converted audio inputs to text and fused them with the text information to obtain a fusion representation. They optimized the common retrieval space using triplet loss, semantic loss, and consistency loss. Experimental findings on multiple datasets indicate the effectiveness of their approach. 
	
	\begin{figure*}
		\centerline{\includegraphics[width=1.7\columnwidth]{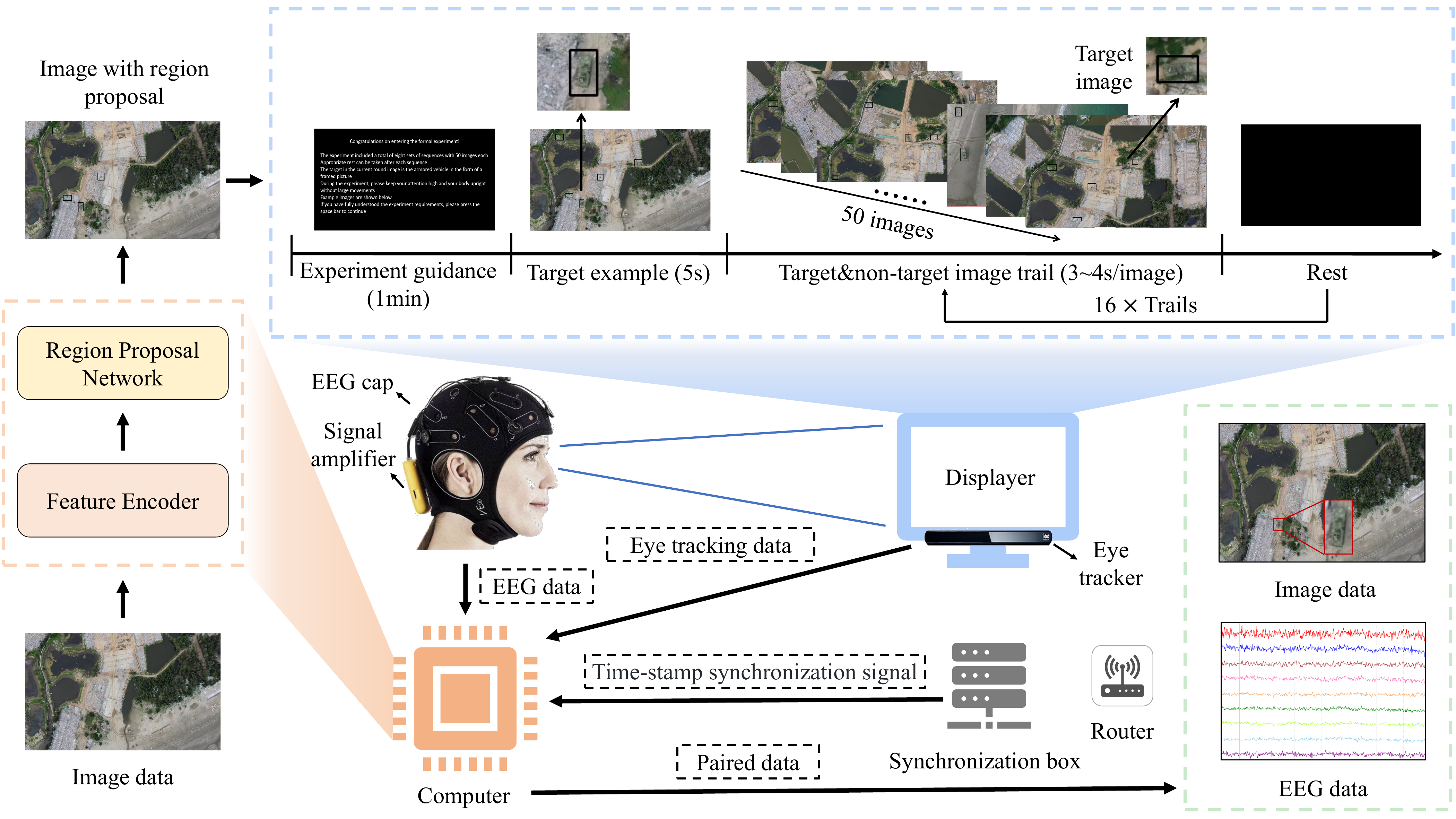}}
		\caption{The data acquisition process of the brain-eye-computer based object detection system. First, the captured image data is pre-processed by the feature encoder and region proposal network to obtain the image with region proposal. Then, the ESSVP paradigm is constructed to elicit the ERPs in EEG signals. After that, the EEG cap collects the EEG signals and sends them to the computer through the signal amplifier. At last, the EEG and image data are combined as the paired data with the help of the eye tracking data and time-stamp synchronization signal from the eye tracker and the synchronization box.}
		\label{figurelabel4}
		\vspace{-1em}
	\end{figure*}
	
	However, these methods focus on feature fusion and loss function design when developing multimodal fusion algorithms, neglecting the potential benefits of knowledge transfer and mutual learning between different modalities. In this paper, we propose a dynamic hybrid fusion approach that integrates multimodal features and trains all branches in the model using an online knowledge distillation (OKD) method. This mutual learning approach for multimodal fusion leverages the visual and cognitive domains to effectively extract crucial information.
	
	\subsection{Knowledge Distillation}
	
	Knowledge Distillation (KD) has attracted considerable attention because of its capacity to compress models. This technique involves training a small student model by imitating the output distribution of a large teacher model~\cite{Hinton2015DistillingTK}. KD has been extensively employed in different fields, including computer vision~\cite{chen2022dearkd}, natural language processing~\cite{ryu2022knowledge}, and emotion recognition~\cite{li2023decoupled}. These approaches can be classified into offline distillation and online distillation according to the distillation schemes~\cite{gou2021knowledge}. Offline distillation typically follows a two-stage training process where a trained teacher model guides the training of a student model. However, this approach is expensive and knowledge transfer is unidirectional. In contrast, online approaches enable end-to-end training, allowing for simultaneous learning of teacher and student models while reducing training~\cite{zhang2018deep}.
	
	In the field of multimodal learning, attention has been drawn to offline knowledge distillation due to its knowledge transfer capabilities. Zhang \emph{et al.}~\cite{Zhang2022VisualtoEEGCK} proposed a visual-to-EEG cross-modal KD method that enhances continuous EEG prediction using dark knowledge from visual modality. Multi-teacher KD methods, including AvgMKD~\cite{you2017learning}, CA-MKD~\cite{zhang2022confidence}, and EMKD~\cite{kwon2020adaptive}, have been proposed to combine knowledge from various modalities and train the student model.
	
	Current online knowledge distillation approaches have achieved significant progress with multimodal homogeneous data. For instance, Zhang \emph{et al.}~\cite{zhang2018deep} introduced DML, demonstrating that student models can learn from each other through their predictions in deep mutual learning. Li \emph{et al.}~\cite{li2023embedded} proposed an innovative embedded OKD approach to leverage ensemble information, overall feature representations from peer networks, and logits to fully exploit the potential of networks. Gou \emph{et al.}~\cite{9926046} proposed CKD-MKT, a collaborative knowledge distillation framework that integrates self-learning and online distillation for multitype knowledge transfer, enabling students to learn from both individual instances and instance relations with two peer networks.
	
	However, offline distillation follows a one-way training paradigm, resulting in high computational costs. Most existing OKD methods are designed for homogeneous data and lack targeted fusion strategies for heterogeneous data. To address this limitation, we propose the AMBOKD method, which effectively extracts features from both 2D EEG data and 3D image data while treating the fusion model's output features output as a single modality for mutual learning. Our method adaptively adjusts the influence weights and optimization levels of each modality, leading to more effective parameters optimization and superior fusion model performance.
	
	\section{System Design and Data Collection}
	\label{seq3}
	In this section, we start by introducing the structure and data acquisition process of the brain-eye-computer based object detection system. Then, we introduce the designed ESSVP paradigm and its experimental environment in detail. Lastly, we present the preprocessing process and the details of the ESSVP dataset. 
	\subsection{System Design}
	
	To address dim object detection in aerial images under few-shot conditions, we propose a brain-eye-computer based object detection system. As shown in Fig.~\ref{figurelabel4}, the system is designed according to EEG, vision, and eye movement properties. Firstly, the computer processes the image data through the feature extractor and the region proposal network (RPN) to obtain the image with the pre-detection box (see Section~\ref{seq32}). Subsequently, this part of the image is presented on the display screen through the eye-tracking-based slow serial visual presentation (ESSVP) paradigm (see Section~\ref{seq33}), which in turn induces the subject to generate the corresponding EEG signals. Through the real-time eye movement data recorded by the eye tracker and the time-stamp synchronization signal of the trigger box, the computer simultaneously extracts the EEG data recorded by the EEG cap and the image of the subject's attention area at the corresponding time. The paired data is then preprocessed and analyzed using the proposed AMBOKD method (see Section~\ref{seq4}). Finally, using eye movement data, the target’s position in the original image is identified to achieve target detection.
	
	
	\begin{figure}
		\centerline{\includegraphics[width=0.8\columnwidth]{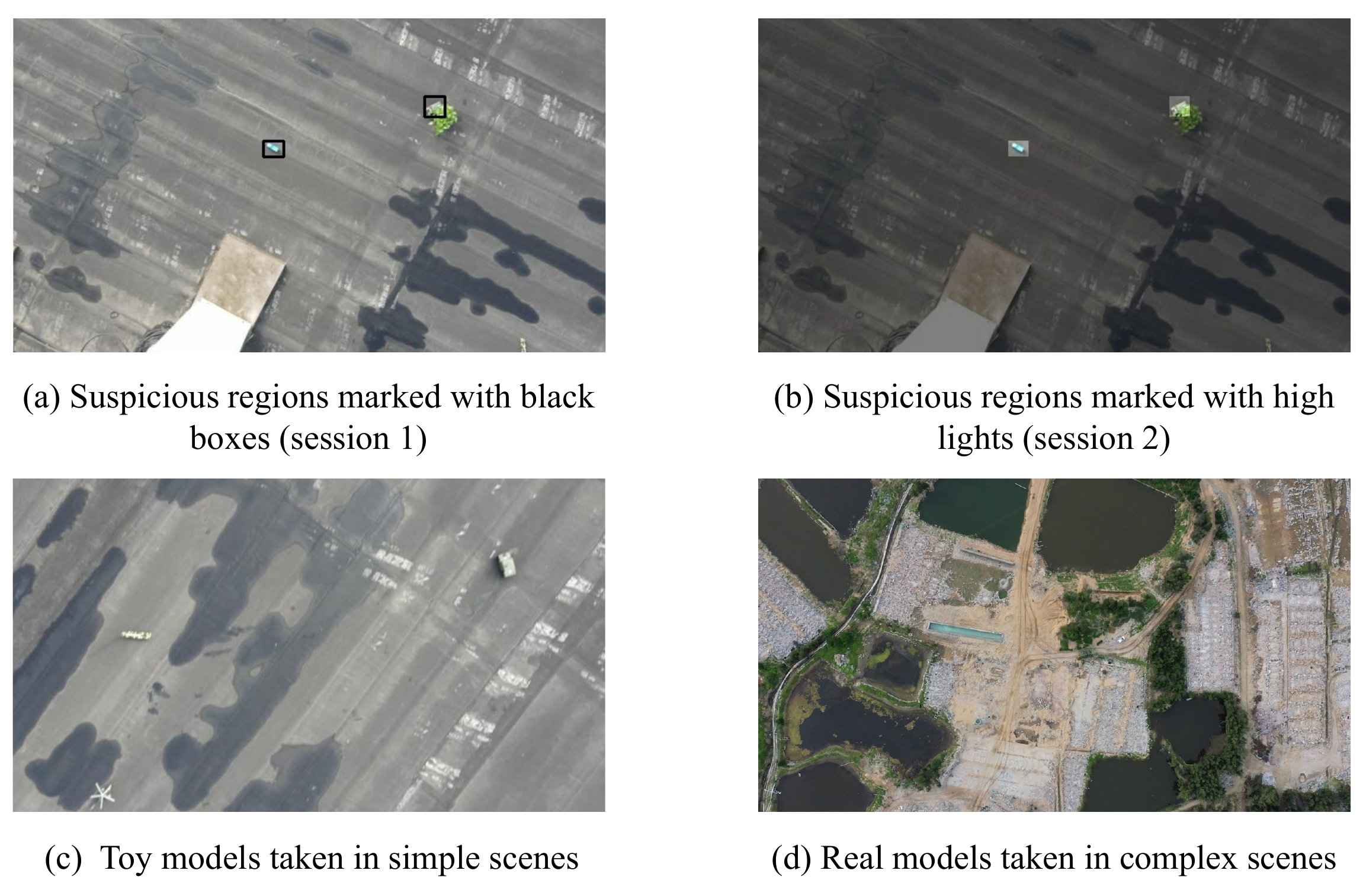}}
		\caption{Examples of stimulus materials.}
		\label{figurelabel5-1}
		\vspace{-2em}
	\end{figure}
	
	\subsection{Suspicious Region Detection}
	
	\label{seq32}
	The RPN is an integral component in deep learning and computer vision, particularly in object detection tasks~\cite{ren2015faster}. Operating by processing a feature map, the RPN employs a sliding window approach across various locations, scales, and aspect ratios using an anchor box. This network evaluates the likelihood of each anchor enclosing an object and adjusts its position and dimensions accordingly, generating precise region proposals. The designed system uses a pre-trained ResNet-50 feature encoder~\cite{he2016deep} to obtain feature maps of image data, and employs RPN to generate region proposals, which are denoted with black boxes. The anchors used have scales of 16, 32, 48, 64, and 80, with aspect ratios of 0.5, 0.75, 1.0, 1.5, and 2.0.
	
	\subsection{ESSVP Paradigm} 
	\label{seq33}
	To elicit ERPs more efficiently and obtain the EEG-image data, we propose the ESSVP paradigm according to the RSVP~\cite{zhang2020benchmark} and the AVEP~\cite{fan2022dc} experimental paradigm. Compared to the RSVP paradigm, the ESSVP paradigm offers improved processing of complex UAV-captured imagery, enabling detection of dim targets rather than conspicuous ones. In this paradigm, visual stimuli are presented at a slow and controlled pace, providing the subjects sufficient time to search for specific targets. By integrating eye movement technology, we can accurately track the observer's gaze and determine the specific areas of interest during the dim target recognition task.
	
	\begin{figure}
		\centerline{\includegraphics[width=0.9\columnwidth]{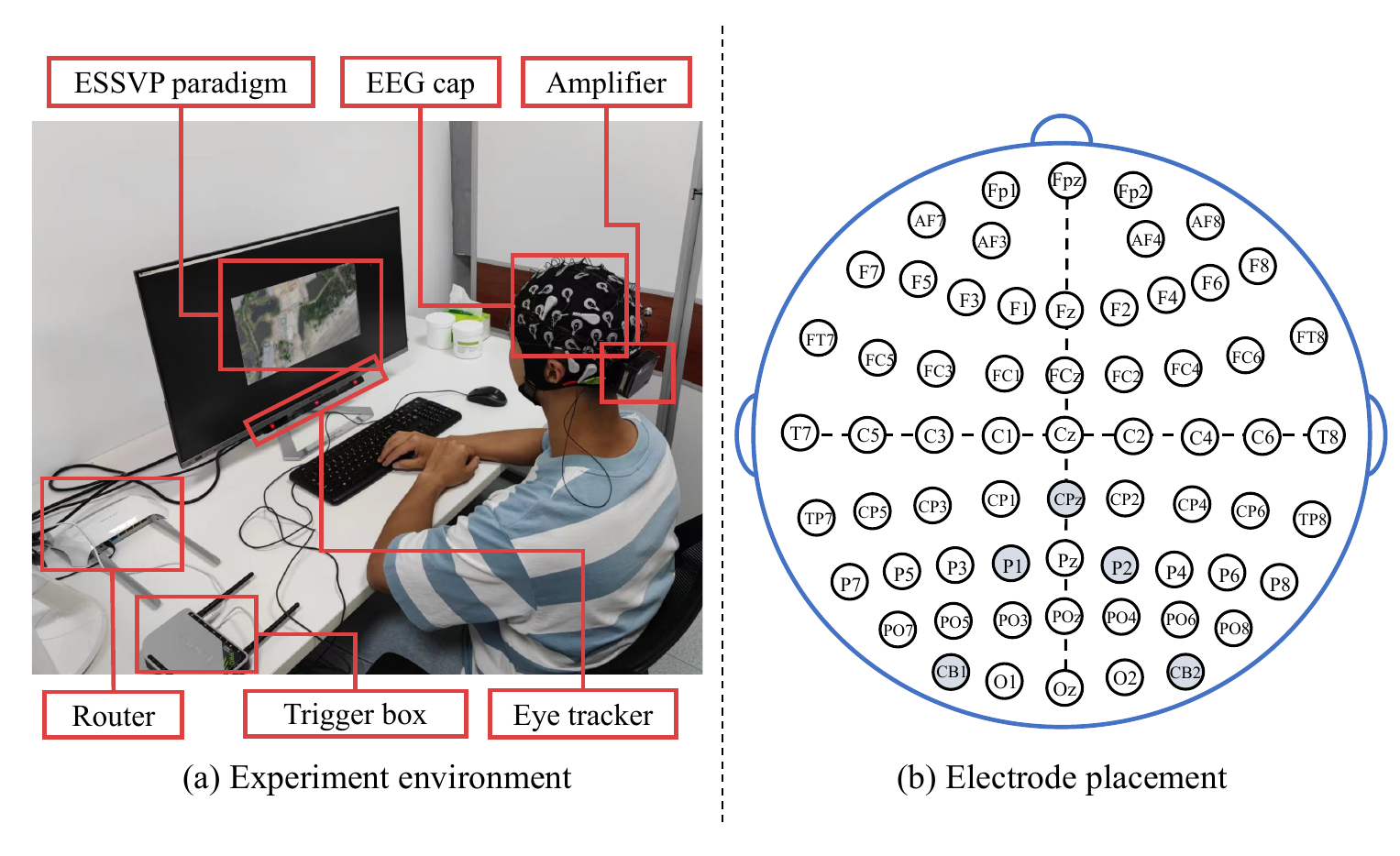}}
		\caption{The experimental configuration of ESSVP.}
		\label{figurelabel5}
		\vspace{-1.5em}
	\end{figure}
	
	The ESSVP paradigm first presents experimental guidance for 1 minute, followed by displaying the target example to the subject for 3–4 seconds, as shown in Fig.~\ref{figurelabel4}. In this formal experiment, participants view 16 sequences of stimuli, each containing 50 images. These sequences are evenly divided into two sessions according to the region marking method. In each session, the first six sequences display images of toy models such as armored vehicles and airplanes, taken in simple scenes, with each image shown for 3\,s. The last two sequences in each session present real images of real models, taken in complex scenes, with each image displayed for 4\,s. The target in the first six sequences is the armored model, whereas in the last two sequences, the target is the vehicle. The different types of images used in the ESSVP paradigm are illustrated in Fig.~\ref{figurelabel5-1}.
	
	According to the RSVP researches~\cite{zhang2020benchmark,cecotti2011impact}, the lower the probability of the target image, the better the detection performance. Since our target images also contain some non-targets and real-world scenarios often involve frequent occurrences of targets, we increased the probability of the target image to 40\% to validate the practicality of our proposed system and method. 1$\sim$2 nontarget images are randomly inserted between each target image to avoid the attentional blink caused by successive targets. During the stimulus presentation, we instruct subjects to actively search for the target among the candidate regions in the image, which is generated by the region proposal network in advance. Each candidate region must be fixated for 0.5\,s. To prevent excessive visual fatigue due to prolonged task engagement, subjects are allowed to rest after each sequence, and the duration of the rest is determined by the subjects themselves.
	
	The eye tracker continuously records the eye movement data of the subjects throughout the experiment. Specifically, the candidate box and image sequence data will be recorded if the subject fixates on a candidate area for more than 0.3s. In addition, to record the EEG signals of subjects during the corresponding time, the fixation event will be sent to the EEG signal acquisition system. Finally, the EEG-image data for the dim object task is generated collecting the EEG signals captured during the fixation periods and the image data from the subject's attention area.
	
	The experimental environment for ESSVP is illustrated in Fig.~\ref{figurelabel5}. Specifically, the EEG signals sampled at a rate of 1000\,Hz are collected through 64 wet electrodes adhering to the 10$\sim$20 standard~\cite{chatrian1985ten} on the EEG cap. The impedance of these electrodes is maintained below 10k. Throughout the experiment, subjects are seated comfortably in a quiet environment, positioned approximately 60\,cm away from the display screen, facing the displayed image. To ensure signal quality, we instruct participants to remain stable and minimize body and head movements during the visual presentation paradigm. This directive aims to minimize potential noise interference to the EEG signals.
	
	\subsection{Data Collection and Preprocessing} 
	
	The ESSVP dataset includes EEG data from 10 subjects (6 males, 4 females) recruited through stratified sampling to ensure demographic diversity. All subjects were right-handed college students aged 22-26 (mean=23.4, SD=1.2) with normal or corrected-to-normal vision and no history of mental illness. All subjects were naive to EEG paradigms and were provided with a clear understanding of the experimental procedure, task requirements, and brief information about the target characteristics before the experiment. In addition, the subjects signed an informed consent form. Ethical approval for the experiment was obtained from the relevant committee.
	
\begin{figure}
		\centerline{\includegraphics[width=0.9\columnwidth]{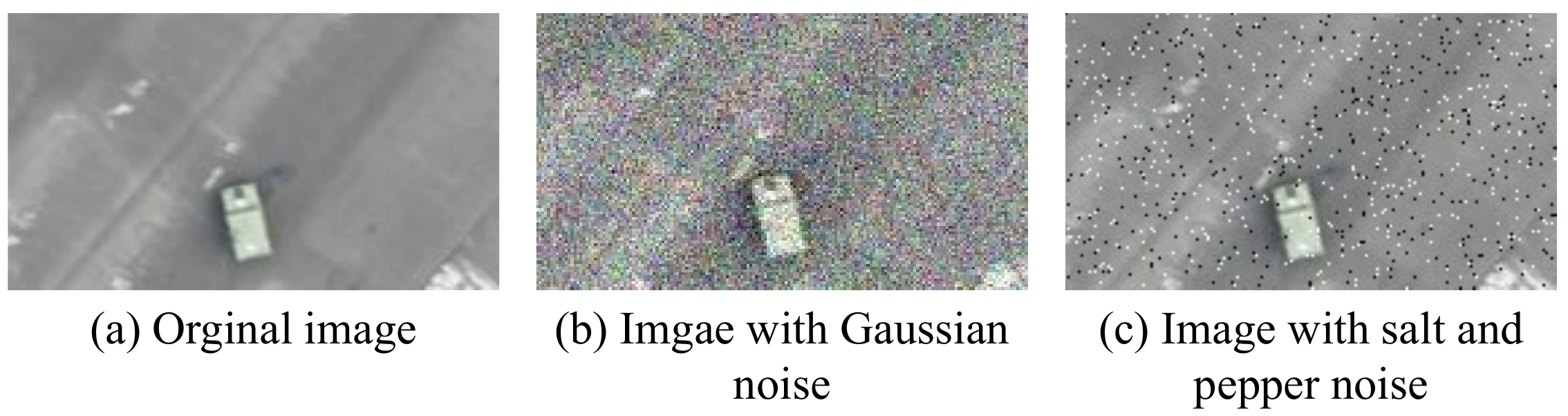}}
		\caption{The noise effect on image data of validation set.}
		\label{figurelabel33}
		\vspace{-1.5em}
	\end{figure}

	This dataset consists of 13405 samples, including 3880 positive samples and 9525 negative samples, with a target probability of 28.9\%. Each sample includes EEG data from 59 electrode channels, as 5 out of the 64 electrodes are redundant and therefore excluded. The EEG data are filtered using a 2$\sim$30\,Hz bandpass filter and downsampled from the original 1000\,Hz to a rate of 250\,Hz. The data is then segmented into 1.2\,s (-500$\sim$700\,ms) samples based on the trigger. Baseline correction is applied by employing the data from the first 200\,ms interval as a reference to subtract the mean activity level. The image data used has a minimum pixel size of $150\times150$, corresponding to the center of the focused candidate box. All images are resized to $224\times224$ pixels to standardize the input for the image network. The images in the validation set are preprocessed by introducing noise to simulate real-world conditions and evaluate the algorithm's robustness, as shown in Fig.~\ref{figurelabel33}. Specifically, one-third of the images are augmented with 0.2 Gaussian noise, and another third with 0.2 salt-and-pepper noise.
	
	The EEG data are represented as a set of labeled samples $\left\{\left(M_{e}^i \in R^{59 \times 300}\right) \mid i=1,2, \ldots, N\right\}$ after the preprocessing steps. Each sample comprises a matrix with dimensions 59 (number of electrode channels) by 300 (number of time points), where the time points correspond to the segmented 1.2\,s intervals that have undergone the bandpass filtering and baseline correction steps.
	
	\begin{figure*}
		\centerline{\includegraphics[width=0.9\textwidth]{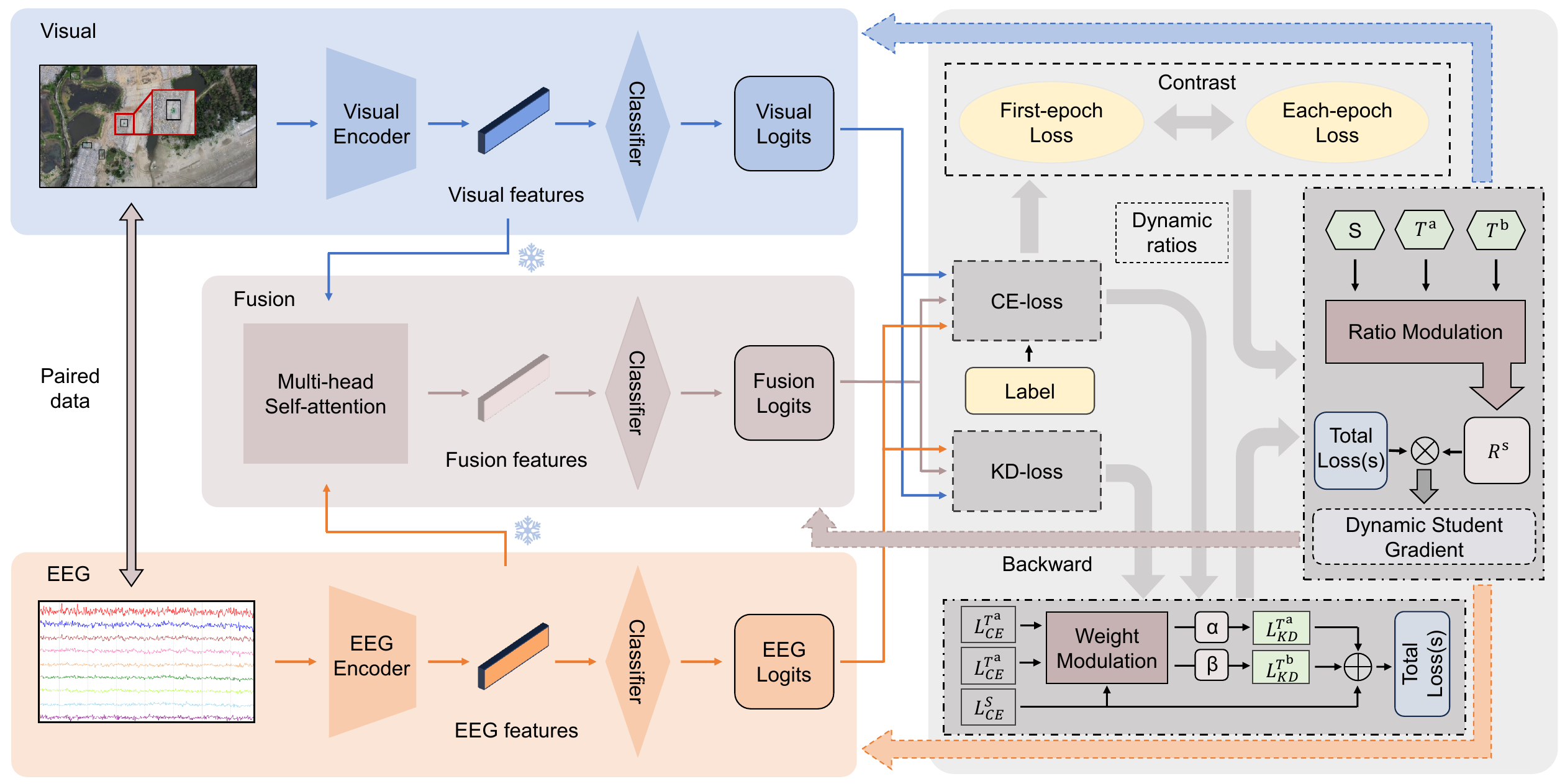}}
		\caption{Overview of the AMBOKD method for dim object recognition and multimodal data processing. First, the \textit{visual encoder} and \textit{EEG encoder} are used to extract preliminary representations from the two domains. Following this, the fusion domain incorporates the \textit{multi-head self-attention} module to process the features from both domains. Subsequently, logits are calculated from three domains, facilitating the computation of the total loss and dynamic learning progress ratios for each modality. Finally, the gradients of each modality are dynamically modulated by the other two modalities and the truth label.}
		\label{figurelabel1}
		\vspace{-1em}
	\end{figure*}
	
	The image data are represented as a set of samples $\left\{\left(M_{v}^{i} \in \mathbb{R}^{3 \times 224 \times 224}\right) \mid i=1,2, \ldots, N\right\}$. Each sample comprises a tensor with dimensions 3 (number of color channels) by 224 (height) by 224 (width), corresponding to the size of the input images in the dataset.
	
	Both the labeled EEG and image modality samples were collected as paired samples, forming sample pairs $X=\left\{\left(M_{v}^{i}, M_{e}^i, y_i\right) \mid i=1,2, \ldots, N\right\}$ in the constructed ESSVP dataset, where $N$ is the total number of paired data, and $y_i$ is the corresponding ground truth label. These sample pairs serve as the foundation for the analysis and assessment of the proposed approach for dim target recognition in aerial images.

	\section{Methodology}
	\label{seq4}
	
	The framework of the proposed adaptive modality balanced online knowledge distillation (AMBOKD) method is illustrated in Fig.~\ref{figurelabel1}. In this method, we use two feature extractors and a multi-head self-attention mechanism (see Sections~\ref{seq41} and~\ref{seq42}) to process EEG and visual data. Then, we introduce the mutual learning with online knowledge distillation and adaptive modality balancing strategies to optimize the fusion model for better performance. (Section~\ref{seq43} and Section~\ref{seq44})
	
	Initially, AMBOKD receives paired data $X$ and extracts EEG and visual features through the \textit{visual encoder} and \textit{EEG encoder}. These features are subsequently passed through the \textit{multi-head self-attention} module to generate the fusion features, and the features from each domain are sent to their respective classifiers. The features extracted from the visual, EEG, and fusion domains are classified and employed to compute the losses of the three modalities using the cross-entropy (CE) and KD loss functions. Subsequently, the three modalities take turns as students and teachers to compute the total loss, the weight of the teacher influence, and the dynamic student gradient for parameter updating. The processes of the feature extraction, feature fusion, the online knowledge distillation, and the adaptive modality balancing are detailed as follows.
	
	\subsection{Extraction of Modality Representation}
	\label{seq41}
	The EfficientNet architecture~\cite{tan2019efficientnet} is employed as the \textit{visual encoder} and MCGRAM~\cite{li2022mcgram} as the \textit{EEG encoder} to extract the specific information of the visual images and EEG signals in dim target recognition. The focus of this initial extraction step is to capture the pertinent information from each modality effectively. The proposed approach maximizes the extraction of valuable information from visual images and EEG signals by designing dedicated networks for each modality. In addition, the logits obtained from the EEG and visual domains denote the independent recognition capability of the uni-encoder model.
	
	\subsubsection{Visual Encoder}
	EfficientNet~\cite{tan2019efficientnet}, a well-designed convolutional neural network, has demonstrated outstanding advancements in accuracy and computational efficiency. This achievement is attributed to the effective balance of network depth, width, and resolution, resulting in enhanced recognition performance. In this study, EfficientNet-B0 was used because of its high efficiency, achieving comparable accuracy to ResNet-50 while requiring fewer parameters and floating point operations for the same input size.
	
	The parameters of EfficientNet in AMBOKD were initialized using pre-training on the ImageNet dataset to ensure optimal performance and efficiency. This enables the leverage of knowledge gained from the ImageNet dataset for improved results in our specific task.
	
	The effective representation of visual images from the visual encoder can be easily extracted with the help of EefficientNet. The representation of the visual domain is computed as
	\begin{equation}
		F_{v}={\operatorname{EfficientNet}}\left({M_{v}}\right).
		\label{eq1}
	\end{equation}
	
	\subsubsection{EEG Encoder}
	MCGRAM~\cite{li2022mcgram} is a compact convolutional neural network specifically designed for EEG signals, which comprises three primary components: the frequency encoder, spatial encoder, and temporal encoder.
	
	The frequency encoder can effectively learn frequency-related features and capture the spectral characteristics of EEG signals by employing specific kernels in the multi-scale convolution module. Spatial representations are learned by the spatial encoder using a graph convolution module, which exploits the inherent spatial relationships between various electrode channels. To extract temporal features and obtain the final global representation of the EEG modality, the temporal encoder uses a two-layer long short-term memory network and a self-attention module to model the temporal dependencies and crucial temporal patterns. 
	
	In summary, MCGRAM captures and encodes relevant information from EEG signals by cascading frequency, spatial, and temporal blocks. This resulted in a comprehensive and informative representation for further analysis and classification tasks. The representation of the EEG modality is computed as
	\begin{equation}
		F_{e}={\operatorname{MCGRAM}}\left({M_{e}}\right).
		\label{eq2}
	\end{equation}
	
	\subsection{Fusion of Modality Representation}
	\label{seq42}
	We develop a fusion model to integrate EEG and visual modality information. This model is considered as a novel modality that collaborates with the original modality in the mutual learning process. To focus on the most important information across various representations subspaces~\cite{vaswani2017attention}, a multihead self-attention module is employed within the fusion model. This module creates a global representation capturing the combined knowledge.
	
	The \textit{multi-head self-attention} module is designed to extract and align crucial information from EEG and visual modality features. Dynamically learned weights are used to weigh and fuse the features of both modalities to obtain global fusion features. Specifically, a fully connected (FC) layer is used to align the EEG feature $F_{e}\in \mathbb{R}^{f^{e}}$ and visual feature $F_{v}\in \mathbb{R}^{f^{v}}$ of each pair of data. To obtain the preliminary fused feature $F_{c}\in \mathbb{R}^{f^{c}}$, the aligned features are then concatenated. Three sets of FC layers were used to independently transform the features $F_{c}$ into queries $Q$, keys $K$, and values $V$ for the attention layer of the $j$-th head. The softmax function is used to multiply, scale, and normalize $Q$ and $K$ to learn the attention scores $A\in \mathbb{R}^{f^{c} \times f^{c}}$ between the two modalities:
	
	\begin{equation}
		{A}=\operatorname{softmax}\left(\frac{Q K^{\top}}{\sqrt{d_l}}\right),
		\label{eq3}
	\end{equation}
	where $\sqrt{d_{l}}$ is employed to scale the matrix multiplication result. Performing matrix multiplication between the values $V$ and the attention scores $A$ describes the output of the single-head attention as follows:
	\begin{equation}
		{H}_j=A_j V_j.
		\label{eq4}
	\end{equation}
	Finally, to obtain the final fusion features $F_{f}$, the outputs of all heads are concatenated and passed through an FC layer with a softmax function:
	\begin{equation}
		F_{f}=\operatorname{softmax}\left(\operatorname{FC}\left({||}_{j}^{J}\left(H_1,H_2,\dots,H_J\right)\right)\right),
		\label{eq5}
	\end{equation}
	where $||$ denotes the concat function and $J$ represents the number of heads.
	
	The features obtained from the three domains are classified separately using their respective classifiers. For example, considering the fusion domain, the classification process can be described as follows:
	\begin{equation}
		\text {G}_{f}=W_{f}{F_{c}}+b_{f}, 
		\label{eq6}
	\end{equation}
	where $W_{f}$ denotes the weights and $b_{f}$ is the biases.
	
	\subsection{Mutual Learning with Online Knowledge Distillation}
	\label{seq43}
	\begin{figure}
		\centerline{\includegraphics[width=0.8\columnwidth]{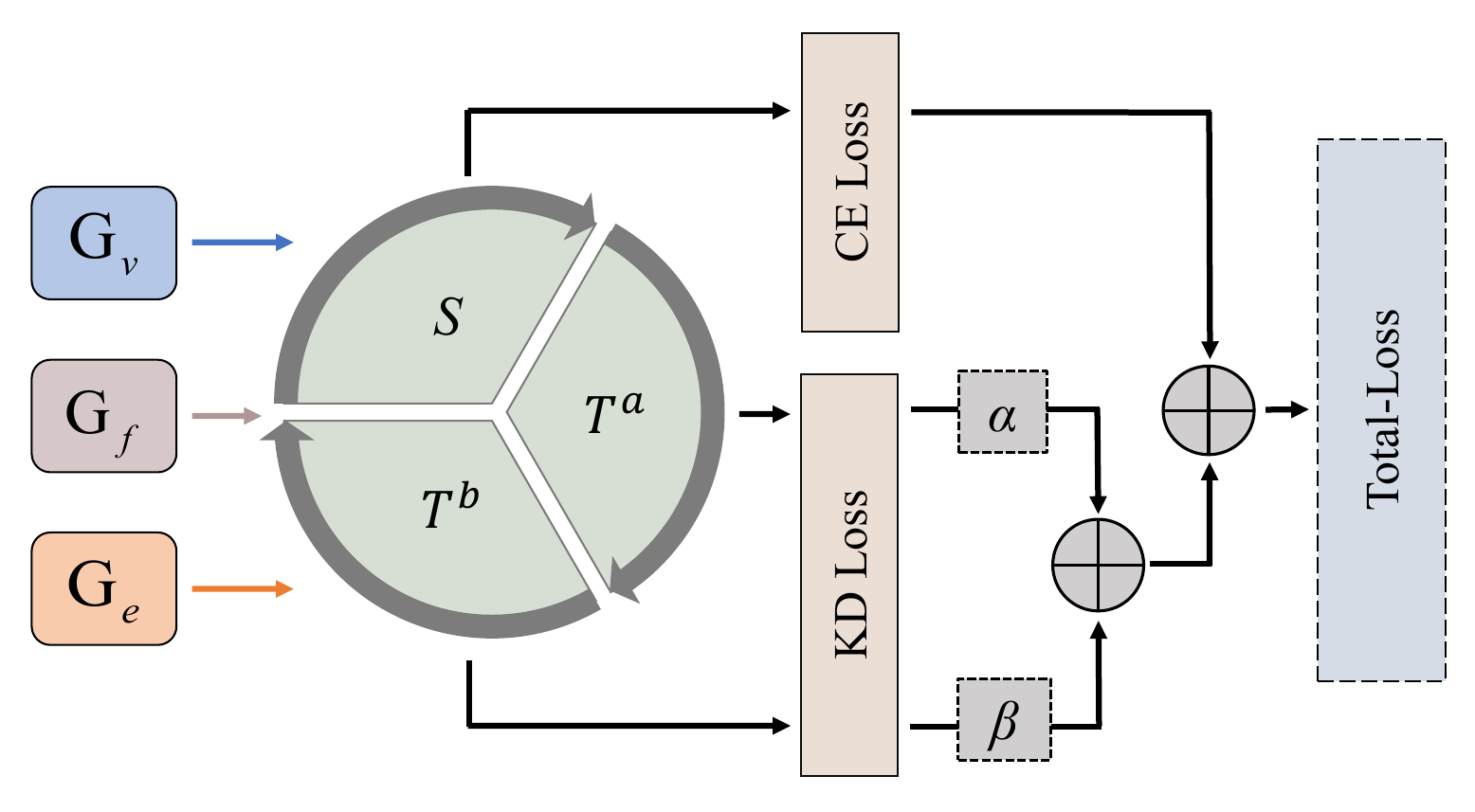}}
		\caption{Strategy of mutual learning with online knowledge distillation (OKD).}
		\label{figurelabel3}
		\vspace{-1em}
	\end{figure}
	
	The OKD approach enables the teacher model and the student model to influence and adapt to each other during the training process. This dynamic interaction allows the teacher model to adjust the knowledge it transfers, thereby improving the student model's performance. By employing the OKD approach to facilitate mutual learning between the modalities, different modalities can learn from each other and progress together throughout the training process, fully leveraging the strengths and improving the performance of each modality. Through iterative optimization of each modality, the optimal fusion model for the entire framework is obtained by incorporating the knowledge from all three modalities.
	
	The training process starts by inputting the logits of the three domains, as illustrated in Fig.~\ref{figurelabel3}. Each modality is then treated as a student model sequentially. For each student modality, its CE loss $L_{CE}^{S}$ and the KD losses $L_{KD}^{T_{a}}$ and $L_{KD}^{T_{b}}$ from two teacher modalities are calculated. The CE loss ($L_{CE}^{S}$) effectively quantifies the discrepancy between the predictions of the student model and the true labels, which is computed as follows:
	
	\begin{equation}
		\mathcal{L}_{CE}^{S}=-\frac{1}{N} \sum_i\left[y_i \log \left(p_i\right)+\left(1-y_i\right) \log \left(1-p_i\right)\right],
		\label{eq7}
	\end{equation}
	where $N$ is the sample size of a batch, and $p_i$ represents the probability assigned by the student model for sample $i$ belonging to the positive class. Although CE loss provides valuable learning signals, it is insufficient for enabling various modalities to learn from each other. The KD loss function effectively quantifies the dissimilarity between the teacher and student models to address this limitation, thereby facilitating the distillation of knowledge from the teacher to the student. The KD loss is computed using the Kullback-Leibler (KL) divergences, given the teacher model:
	\begin{equation}
		\mathcal{L}_{KD}^{T}=\frac{1}{N} \sum_i q_i \log \frac{q_i}{p_i},
		\label{eq8}
	\end{equation}
	where $q_i$ denotes the probability assigned by the teacher model for sample $i$ belonging to the positive class.
	
	By considering the CE loss and the KD loss, the overall loss $\mathcal{L}_{total}$ for each modality is obtained as follows:
	\begin{equation}
		\mathcal{L}_{total}= \mathcal{L}_{CE}^{S} + \alpha {\tau}^2 \mathcal{L}_{KD}^{T_{a}} + \beta {\tau}^2 \mathcal{L}_{KD}^{T_{b}},
		\label{eq9}
	\end{equation}
	where $\tau$ is the distillation temperature in knowledge distillation. $\alpha$ and $\beta$ represent the interaction weights that determine the influence contribution of distillation from the two teacher models and balance the magnitude difference between the CE loss and KD loss.
	
	\subsection{Mutual Learning with Adaptive Modality Balancing}
	\label{seq44}
	To address imbalance optimization~\cite{peng2022balanced} in the learning process of each modality, an AMB module for mutual learning with OKD is proposed. There are two blocks in the module: dynamic weights for KD losses and dynamic ratios for backward gradients, detailed as follows.
	
	\subsubsection{Dynamic Weights for KD Losses}
	
	In the online knowledge distillation training process, the KD loss from the teacher model encapsulates the distilled knowledge. Most existing methods~\cite{you2017learning,yang2023online,li2023embedded} compute the total training loss by adding the cross-entropy loss to the knowledge distillation loss, typically setting $\alpha$ and $\beta$ to 1 in Eq.~\eqref{eq9}. However, due to modality differences, simply averaging fails to effectively capture the superior knowledge from the teacher model. Furthermore, manually setting a fixed distillation loss weight often entails extensive experimentation and significant time costs. 
	
	To address the challenges mentioned above, we introduce a weight modulation block that dynamically calculate the weights for KD losses from different teacher models. This block efficiently extract knowledge during multimodal mutual learning. The entire workflow is described in Algorithm~\ref{alg1} in detail. The saturation function $\text{Sat}\left(\cdot\right)$ is defined as follows for $a<b$: $\text{Sat}\left(x,a,b\right)=a$ if $x \geq a$, $\text{Sat}\left(x,a,b\right)=x$ if $a \geq x \geq b$, $\text{Sat}\left(x,a,b\right)=b$ if $x \leq b$. The parameters $\alpha_{\min}$ and $\alpha_{\max}$ denote the lower and upper bounds of $\alpha$, similar to $\beta_{\min}$ and $\beta_{\max}$ for $\beta$. 
	
	\subsubsection{Dynamic Ratios for Backward Gradients}
	Current multimodal methods lack the consideration of the modality imbalance optimization. Thus, we propose a dynamic gradient modulation block to dynamically balance the modal optimization level in multimodal mutual learning process, ensuring full optimization of different modalities. 
	
	\begin{algorithm}[t]
		\small
		\setstretch{1.15}
		\caption{The parameters updating process}
		\begin{algorithmic}[1]
			\REQUIRE Training Dataset $X$, iteration number $T$, batchsize $B_m$, model parameters $\theta^{S}$, learning rate $\eta$, hyper-parameter $\gamma$, current epoch $E_n$
			
			\FOR{$t=0, 1, \ldots, T$}
			\STATE Sample minibatch $B_{t}$ from $X$ with batchsize$B_m$
			\STATE Feed-forward the batched data $B_t$ to the model\\
			\vspace{2mm}
			{/** \textit{Dynamic weights modulation} **/}
			\vspace{2mm}
			\STATE Compute the CE losses $\mathcal{L}_{CE_{(t)}}^{S}$, $\mathcal{L}_{CE_{(t)}}^{T_{a}}$ and $\mathcal{L}_{CE_{(t)}}^{T_{b}}$
			\STATE Compute the KD losses $\mathcal{L}_{KD_{(t)}}^{T_{a}}$ and $\mathcal{L}_{KD_{(t)}}^{T_{b}}$
			\STATE Compute $\alpha = \text{Sat}\Big(\frac{\mathcal{L}_{CE_{(t)}}^{S}}{\mathcal{L}_{CE_{(t)}}^{T_{a}}}, {\alpha}_{\min}, {\alpha}_{\max}\Big)$
			\STATE Compute $\beta = \text{Sat}\Big(\frac{\mathcal{L}_{CE_{(t)}}^{S}}{\mathcal{L}_{CE_{(t)}}^{T_{b}}}, {\beta}_{\min}, {\beta}_{\max}\Big)$
			\STATE Compute the total loss $\mathcal{L}_{{total}_{(t)}}^{S} = \mathcal{L}_{CE_{(t)}}^{S} + \alpha {\tau^2} \mathcal{L}_{KD_{(t)}}^{T_{a}} +\beta {\tau^2} \mathcal{L}_{KD_{(t)}}^{T_{b}}$\\
			\vspace{2mm}
			{/** \textit{Dynamic gradients modulation} **/}
			\vspace{2mm}
			\IF{$E_n == 1$}
			\STATE Calculate average CE losses $\mathcal{L}_{\rm{base}}^{S}, \mathcal{L}_{\rm{base}}^{T_{a}}, \mathcal{L}_{\rm{base}}^{T_{b}}$
			\STATE Set $R_{(t)}^{DG} \leftarrow 1$
			\ELSE
			
			\STATE Update $R^{S}_{(t)} \leftarrow \frac{\mathcal{L}_{\rm{base}}^{S}-\mathcal{L}_{CE_{(t)}}^{S}}{\mathcal{L}_{\rm{base}}^{S}}$, $R_{(t)}^{T_{a}} \leftarrow \frac{\mathcal{L}_{\rm{base}}^{T_{a}}-\mathcal{L}_{CE_{(t)}}^{T_{a}}}{\mathcal{L}_{\rm{base}}^{T_{a}}}$, $R_{(t)}^{T_{b}} \leftarrow \frac{\mathcal{L}_{\rm{base}}^{T_{b}}-\mathcal{L}_{CE_{(t)}}^{T_{b}}}{\mathcal{L}_{\rm{base}}^{T_{b}}}$
			\STATE Compute $R_{(t)}^{DG} = \text{Sat}\Big(\big(\frac{R_{(t)}^{T_{a}}+R_{(t)}^{T_{b}}}{2 \times R_{(t)}^{S}}\big)^{\gamma}, R_{\min}, R_{\max}\Big)$
			\ENDIF
			\STATE Compute $\tilde{g}\left(\theta_{(t)}^S\right)=\frac{1}{B_m} \sum_{x \in B_{(t)}} \nabla_{\theta^{S}} \mathcal{L}_{{total}_{(t)}}^{S}$
			\vspace{1mm}
			\STATE Compute $D^{\rm{mean}}_{(t)}=\beta_1 \cdot D^{\rm{mean}}_{(t-1)}+\left(1-\beta_1\right) \cdot \tilde{g}\big(\theta_{(t)}^S\big)$
			\vspace{1mm}
			\STATE Compute $D^{\rm{var}}_{(t)}=\beta_2 \cdot D^{\rm{var}}_{(t-1)}+\left(1-\beta_2\right) \cdot \tilde{g}^{2}\big(\theta_{(t)}^S\big)$
			\vspace{1mm}
			\STATE Update ${\theta}^{S}_{(t+1)} \leftarrow {\theta}^{S}_{(t)}-R_{(t)}^{DG} \cdot \frac{\eta}{\sqrt{D^{\rm{var}}_{(t)}}+\epsilon} \cdot D^{\rm{mean}}_{(t)}$
			
			\ENDFOR
			\ENSURE ${\theta}^{S}$

		\end{algorithmic}

		\label{alg1}
	\end{algorithm}
	
	We further analysis the imbalance optimization phenomenon during mutual learning by calculating the backward gradient of each modality. In the backward phase of the training process, when the fusion modality acts as the student, the gradient depends on the CE loss can be calculated as $\frac{\partial \mathcal{L}_{CE}^{f}}{\partial G_f^i}$ using the following formulas. The CE loss of the fusion modality, denoted by $\mathcal{L}_{CE}^{f}$, is given by
	\begin{equation}
		\mathcal{L}_{CE}^{f} = -\frac{1}{N} \sum_i^{N} \log \frac{e^{G_{f\left(y_i=c\right)}^i}}{\sum_{k=1}^{M} e^{G_{f\left(y_i=k\right)}^i}},
		\label{eq10}
	\end{equation}
	where $G_{f\left(y_i=c\right)}^{i}$ is the predicted value when the true label $y_i$ equals class $c$, and $M$ is the total number of categories. $\operatorname{G}_{f}^{i}$ is the $i$-th logit output of the fusion model, calculated as
	\begin{equation}
		\operatorname{G}_{f}^{i} = W_f \left( W_f^v \phi_v(\theta_v, F_v^i) + W_f^e \phi_e(\theta_e, F_e^i) \right) + b_f,
		\label{eq11}
	\end{equation}
	where $\phi_v(\theta_v,\cdot)$ and $\phi_e(\theta_e,\cdot)$ represent the visual and EEG modality encoders, respectively, with $\theta_v$ and $\theta_e$ as their parameters. $W_f^e$ and $W_f^v$ are the weight matrices that determine the relative significance of the features extracted from the visual and EEG modalities. $W_f$ and $b_f$ act as the parameters in the fusion process.
	
	Combined with Eq.~\eqref{eq10} and Eq.~\eqref{eq11}, the gradient can be expressed as
	\begin{equation}
		\frac{\partial \mathcal{L}_{CE}^{f}}{\partial G_{f\left(y_i=c\right)}^{i}} = \frac{e^{{\left(W_fW_f^v\phi_v^i+W_fW_f^e\phi_e^i+b_f\right)}_{y_i=c}}}{\sum_{k=1}^{M} e^{{\left(W_fW_f^v\phi_v^i+W_fW_f^e\phi_e^i+b_f\right)}_{y_i=k}}}-1_{y_{i}=c},
		\label{eq12}
	\end{equation}
	$\phi_v(\theta_v,F_v^i)$ and $\phi_e(\theta_e,F_e^i)$ are simplified as $\phi_v^i$ and $\phi_e^i$ for convenience. When a particular modality, such as the visual modality, exhibits classification performance, it contributes more to the gradient $\frac{\partial \mathcal{L}_{CE}^{f}}{\partial G_f^i}$ through the expression $W_fW_f^v\phi_v^i$, thereby leading to a lower loss globally. Therefore, the EEG modality, characterized by lower confidence in accurate predictions, will obtain limited optimization during parameter updates via backpropagation. This phenomenon indicates that unimodal models may become overtrained or underoptimized as the multimodal model approaches convergence.
	
	In this imbalanced training phenomenon, the disparate learning efficiencies of unimodality hinder the performance of the fusion model, resulting in under-optimized and overfitting representations that constrain the overall model performance. Thus, our proposed dynamic modulation block aims to dynamically balance the learning levels of modalities until they reach their optimal performance.
	
	The updating process is presented in Algorithm~\ref{alg1}. Specifically, the ratios $R^{S}_{(t)}$, $R^{T_a}_{(t)}$ and $R^{T_b}_{(t)}$ are computed for indicating the current training optimization level of each modality. To ensure consistent training optimization progress and maximize the completeness and effectiveness of the optimization in multimodal mutual learning, the dynamic learning progress ratio in $t$-th iteration step is computed as
	\begin{equation}
		R_{(t)}^{DG} = 
		\begin{cases} 
			1 & \text{if } E_n = 1 \\
			\text{Sat}\Big(\big(\frac{R_{(t)}^{T_{a}}+R_{(t)}^{T_{b}}}{2 \times R_{(t)}^{S}}\big)^{\gamma}, R_{\min}, R_{\max}\Big) & 
			\text{if } E_n > 1
		\end{cases}
		\label{eq14}
	\end{equation}
	where $R_{\min}$ and $R_{\max}$ represent the lower and upper bounds in the function, $\gamma$ is a hyper-parameter to control the degree of modulation. 
	
	The extensively used adaptive moment estimation (Adam) optimization approach is employed in the backpropagation process. The primary idea underlying the Adam algorithm is to maintain a running average of the first-order moment (mean) and the second-order moment (variance) of the gradients. This facilitates the estimation of adaptive learning rates for various parameters during optimization. Using the dynamic modulation ratio $R_{(t)}^{DG}$, the parameters $\theta^{S}$ in the student model are optimized adaptively based on the modality learning progress. The updating process is expressed as
	
	\begin{equation}
		{\theta}^{S}_{(t+1)} \leftarrow {\theta}^{S}_{(t)}-R_{(t)}^{DG} \frac{\eta}{\sqrt{D^{\rm{var}}_{(t)}}+\epsilon} \cdot D^{\rm{mean}}_{(t)},
		\label{eq15}
	\end{equation}
	where $\eta$ is the learning rate and $\epsilon$ represents a small constant added to the denominator for numerical stability. $D^{\rm{mean}}_{(t)}$ and $D^{\rm{var}}_{(t)}$ represent the exponentially decaying average of the past gradients' first moment estimate (mean) and second moment estimate (variance) at time $t$.
	
	Incorporating the AMB module into the loss computing process and the optimization process addresses the challenge of modality imbalance in the multimodal mutual learning process. This mechanism ensures that all modalities learn the effective knowledge and optimized at a comparable pace during mutual learning. Consequently, enhanced completeness, integration, and efficiency are achieved by the fusion model.
	\begin{table*}	
		\centering
		\renewcommand{\arraystretch}{1.1}
		\caption{The comparison studies of the baseline and state-of-art methods in subject-independent experiments}
		
		\label{tab1}
		\setlength{\tabcolsep}{3pt}
		\resizebox{\linewidth}{!}
		{
			\begin{tabular}{ccccccccccc}
				\toprule
				\multirow{2}*{Method} & \multicolumn{5}{c}{Session 1} & \multicolumn{5}{c}{Session 2} \cr
				\cmidrule(lr){2-6}
				\cmidrule(lr){7-11}
				& AUC/Std (\%) & F1/Std (\%) & ACC/Std (\%) & Precision/Std (\%) & \textit{p}-value (AUC) & AUC/Std (\%) & F1/Std (\%) & ACC/Std (\%) & Precision/Std (\%) & \textit{p}-value (AUC) \cr
				\midrule
				EEGNet & 70.65/6.55 & 69.12/6.07 & 71.68/4.98 & 69.84/6.14 & $<10^{-3}$ & 68.86/5.46 & 67.44/5.76 & 69.51/5.71 & 68.00/5.20 & $<10^{-3}$\cr
				MCGRAM & 82.47/7.11 & 75.18/8.47 & 76.69/6.55 & 76.04/9.48 & $<10^{-3}$ & 80.44/5.32 & 72.33/6.12 & 73.59/5.81 & 74.34/4.95 & $<10^{-3}$ \cr
				AMBOKD-E\textsuperscript{*} & \textbf{83.56}/6.61 & \textbf{77.66}/8.21 & \textbf{78.59}/7.19 & \textbf{78.25}/8.31 & $-$ & \textbf{81.85}/5.71 & \textbf{75.95}/5.75 & \textbf{76.71}/5.09 & \textbf{77.05}/5.07 & $-$ \cr
				\midrule
				ResNet-50 & 84.55/7.54 & 75.45/11.26 & 77.37/9.93 & 79.84/5.21 & $<10^{-3}$ & 84.42/6.63 & 76.04/7.43 & 77.04/7.36 & 79.53/6.03& $<10^{-3}$\cr
				EfficientNet & 78.81/8.89 & 81.82/2.79 & 83.96/2.40 & 87.02/1.47 & $<10^{-3}$ & 84.13/3.92 & 80.68/4.63 & 82.81/4.15 & 86.53/2.34 & $<10^{-3}$\cr
				AMBOKD-V\textsuperscript{*} & \textbf{88.54}/3.50 & \textbf{83.08}/2.52 & \textbf{84.88}/2.25 & \textbf{87.54}/1.49 & $-$ & \textbf{87.88}/5.22 & \textbf{82.22}/3.84 & \textbf{83.96}/3.59 & \textbf{87.18}/2.15 & $-$ \cr
				\midrule
				MKD & 90.21/4.45 & 82.87/2.45 & 84.73/2.21 & 87.54/1.39 & $<10^{-3}$ & 89.35/3.54 & 81.80/3.97 & 83.65/3.68 & 87.04/2.12 & $<10^{-3}$ \cr
				EMKD & 90.51/4.22 & 82.65/2.42 & 84.57/2.18 & 87.42/1.34 & $<10^{-3}$ & 88.93/3.50 & 80.95/4.00 & 83.00/3.68 & 86.62/2.06 & $<10^{-3}$ \cr
				CA-MKD & 90.52/4.19 & 82.78/2.52 & 84.67/2.24 & 87.46/1.42 & $<10^{-3}$ & 89.10/3.39 & 81.45/4.03 & 83.39/3.70 & 86.83/2.20 & $<10^{-3}$ \cr
				C2KD & 87.75/4.13 & 82.33/2.82 & 84.34/2.46 & 87.28/1.52 & $<10^{-3}$ & 86.63/4.35 & 82.08/4.05 & 83.87/3.72 & 87.18/2.15 & $<10^{-3}$ \cr
				DIST+ & 90.39/4.19 & 83.42/5.21 & 84.57/5.15 & 86.93/2.78 & $<10^{-3}$ & 89.77/3.66 & 83.12/4.37 & 84.43/4.27 & 86.92/2.83 & $<10^{-3}$ \cr
				DML\textsuperscript{*} & 90.46/3.68 & 79.73/4.97 & 82.44/3.85 & 85.89/2.38 & $<10^{-3}$ & 88.87/3.21 & 77.66/5.50 & 80.60/4.52 & 84.84/2.87 & $<10^{-3}$ \cr
				KDCL\textsuperscript{*} & 89.91/4.22 & 79.05/4.72 & 81.74/3.71 & 84.55/2.25 & $<10^{-3}$ & 89.12/3.11 & 78.70/4.63 & 81.34/4.02 & 85.42/2.30 & $<10^{-3}$ \cr
				EML\textsuperscript{*} & 89.21/4.02 & 80.25/2.92 & 82.52/2.54 & 84.74/1.92 & $<10^{-3}$ & 90.64/3.06 & 80.89/4.06 & 82.93/3.74 & 86.40/2.16 & $<10^{-3}$ \cr
				MLEMKD\textsuperscript{*} & 89.12/2.02 & 80.32/2.28 & 82.83/2.14 & 86.28/1.24 & $<10^{-3}$ & 88.45/2.04 & 79.31/4.28 & 81.76/3.94 & 85.85/2.12 & $<10^{-3}$ \cr
				AMBOKD (ours)\textsuperscript{*} & \textbf{93.66}/2.95 & \textbf{83.33}/2.51 & \textbf{85.08}/2.24 & \textbf{87.82}/1.38 & $-$ & \textbf{93.30}/2.64 & \textbf{82.32}/3.66 & \textbf{84.05}/3.42 & \textbf{87.27}/2.02 & $-$\cr
				\bottomrule
				
			\end{tabular}
		}
		\label{table0}
		{\raggedright
			\textsuperscript{*} \small Online knowledge distillation method\par}
		
		\vspace{-1em}
	\end{table*}
	
	\section{Experimental Results}
	\label{seq5}	
	In this section, the experimental settings with the parameter configurations of our proposed approach are first presented. Subsequently, a comprehensive comparison and ablation study is conducted to evaluate the performance of the proposed AMBOKD approach in the dim object recognition task. System verification through simulated real-world scenarios is designed to further validate the effectiveness of the system for dim object detection.
	
	\subsection{Experimental Settings}
	\subsubsection{Configurations}
	The parameters of the visual and the EEG encoders are set to follow the configurations as described in the previous study~\cite{tan2019efficientnet}~\cite{li2022mcgram}. The number of heads is set to 2 in the multi-head self-attention module of the fusion model. The kernel sizes of the FC layers for alignment are configured as $1280 \times 64$ and $256 \times 64$ for visual and EEG features, respectively. Before computing the Q, K, and V matrices, the matrices in the FC layer have a size of $64 \times 64$. The FC layers have a size of $128 \times 2$ in the classifier of the Fusion domain.
	
	The temperature coefficient $\tau$ is set to 4 in the OKD approaches to control the softness of the KD targets. The hyperparameters $\alpha$ and $\beta$ are dynamically adjusted by the AMB module, which contributes to the regularization component of the overall loss function. The hyperparameter $\gamma$ is set to 3 to fine-tune the sensitivity of the AMB module. In addition, we impose lower and upper limits, $R_{\min}$ and $R_{\max}$ at 0.1 and 10, respectively, to confine the range of dynamic modulation ratio $R^{DG}_{(t)}$ during the training process.
	
	In the experiments, all models are implemented using the PyTorch framework. The Adam optimizer is employed for parameter optimization with a learning rate of $10^{-4}$. The number of training epochs is set to 15, and the batch size is 64. The optimization loss function consists of cross-entropy (CE) loss and knowledge distillation (KD) loss, with specific configurations varying by model.
	
	To evaluate model performance, we adopt a cross-validation approach. Specifically, for each validation fold, the data from one subject is used as the validation set, while the remaining nine subjects' data serve as the training set. This process is repeated across all subjects, and each fold is further evaluated using five different random seeds. The final results are obtained by averaging the performance across all folds and seeds.
	
	\subsubsection{Performance Metrics}
	A comprehensive set of metrics, including area under the receiver operating characteristic curve (AUC), accuracy (ACC), F1 score (F1) and Precision is used to evaluate the performance of classification models. Each metric is analyzed in terms of mean and standard deviation, offering insight into the reliability of the model and its ability to generalize across various datasets or conditions. Specifically, the AUC is emphasized as the main metric because of its effectiveness in measuring the discriminative ability of a model, especially in scenarios with unbalanced class distributions. To evaluate the object detection performance of the brain-eye-computer based system, AP@50:95 and AR@50:95 are used to measure the average precision and recall over Intersection Over Union (IOU) thresholds ranging from 0.50 to 0.95, while AP@50 and AR@50 are used for an IOU of 0.50. In addition, F1@50:95 and F1@50 provide a balanced view of accuracy by considering both precision and recall at these thresholds. 
	
	Additionally, to determine the statistical significance of performance differences, we first conduct a one-way analysis of variance (ANOVA) to assess significant differences among all models. Since we focus specifically on comparing our proposed model against other baseline models, we further conduct pairwise t-tests between the proposed model and each baseline model. The significance threshold is set at $p<0.05$. To address potential multiple comparison issues, Bonferroni correction is applied to adjust the p-values and control the family-wise error rate.
	\vspace{-1em}
	\subsection{Comparison Results}
	\subsubsection{Comparison Results on ESSVP Dataset}
	\label{seq52}
	To verify the effectiveness of the proposed AMBOKD approach, we conduct subject-independent comparison experiments against baseline and state-of-the-art models. Single-modal EEG methods include EEGNet~\cite{lawhern2018eegnet} and MCGRAM~\cite{li2022mcgram}, while single-modal vision methods include ResNet-50~\cite{he2016deep} and EfficientNet~\cite{tan2019efficientnet}. A brief overview of the compared multimodal methods is as follows:
		\begin{itemize}
			\item \textbf{MKD}~\cite{you2017learning}: A multimodal knowledge distillation approach that transfers knowledge from pre-trained EEG and visual teacher models.
			\item \textbf{EMKD}~\cite{kwon2020adaptive}: An entropy-based adaptive multi-teacher knowledge distillation method.
			\item \textbf{CA-MKD}~\cite{zhang2022confidence}: A multi-teacher knowledge distillation technique that adaptively weights teacher predictions based on sample-wise confidence.
			\item \textbf{C2KD}~\cite{huo2024c2kd}: A cross-modal knowledge distillation method that uses bidirectional distillation, on-the-fly selection to filter misaligned samples, and proxy models to transfer knowledge across modalities.
			\item \textbf{DIST+}~\cite{huang2025dist+}:  A knowledge distillation method using correlation-based loss and semantic similarities, with teacher acclimation and feature-level loss extension.
			\item \textbf{DML}~\cite{zhang2018deep}: An online knowledge distillation method where multiple student models learn from each other by aligning predictive distributions.
			\item \textbf{KDCL}~\cite{guo2020online}: An online knowledge distillation method that integrates contrastive learning to improve feature representation.
			\item \textbf{EML}~\cite{li2023embedded}: A method combining mutual learning and ensemble learning, where student models distill knowledge from both individual peers and an ensemble of predictions.
			\item \textbf{MLEMKD}~\cite{10852525}: A mutual learning-empowered knowledge distillation framework that uses a soft-label evaluation mechanism to mitigate redundancy and improve knowledge transfer.

		\end{itemize}
	In addition, the EEG and visual modal model trained in AMBOKD are named AMBOKD-E and AMBOKD-V. 
	
	\begin{figure*}
		\centerline{\includegraphics[width=1.6\columnwidth]{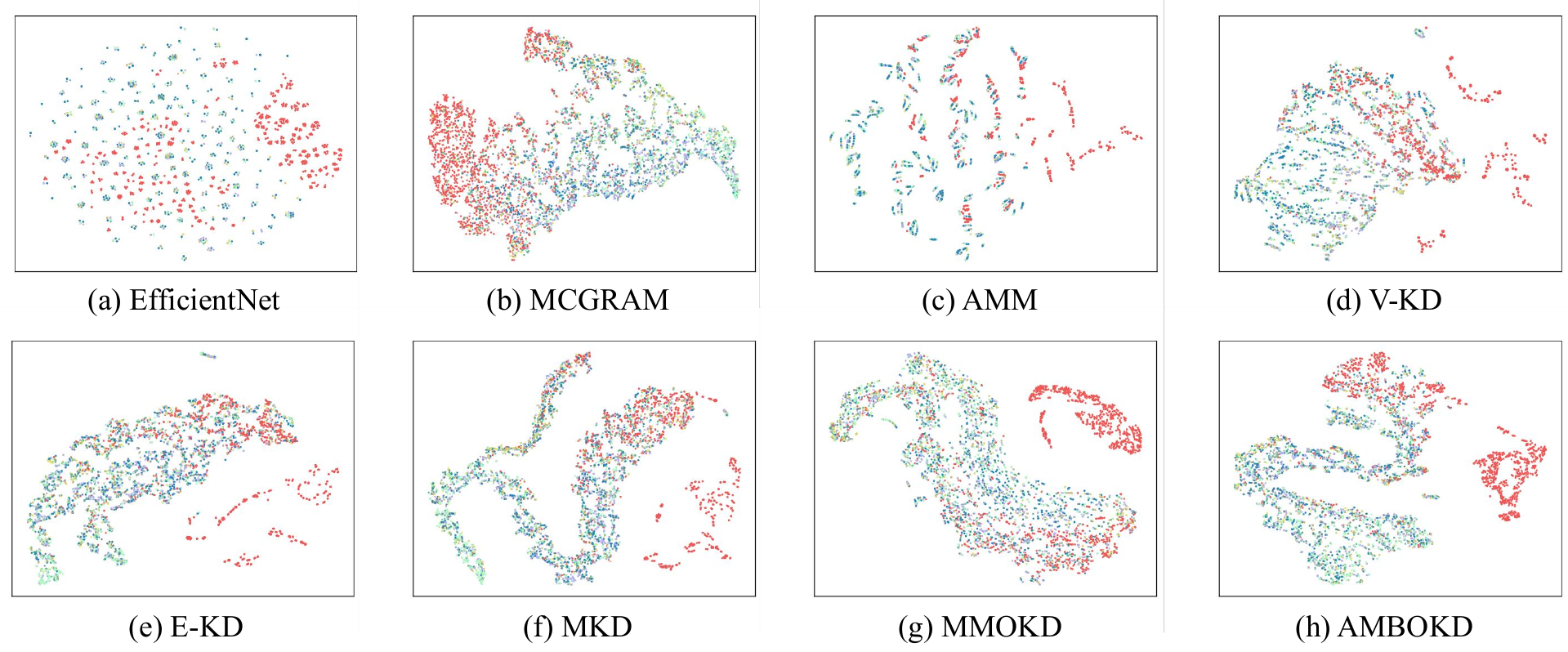}}
		\caption{Visualization of latent representations by the t-SNE algorithm in the ablation studies.}
		\label{figurelabel6}
		\vspace{-1em}
	\end{figure*}
	
	As indicated in Table~\ref{table0}, our proposed AMBOKD achieves impressive results in the dim object recognition task of subject-independent experiments. This outcome surpasses the performance of baseline and state-of-the-art approaches. Additionally, the mutual learning mode with the AMB module enhances the performance of EEG and visual modality models: AMBOKD-E and AMBOKD-V outperform the original unimodal models MCGRAM and EfficientNet across AUC, ACC, and F1 metrics. The AUC-based ANOVA statistical test findings demonstrate that our proposed AMBOKD, AMBOKD-E, and AMBOKD-V significantly outperform all comparison approaches in subject-independent experiments ($p<0.05$).
	
	These results lead to the conclusion that the AMBOKD method effectively combines the cognitive and visual domains, extracting fused global representations. Furthermore, it enables dynamic equilibrium mutual learning between unimodality models and the fusion modality during the training process, leading to improved performance in the fusion, EEG, and visual models.
	

	\subsubsection{Comparison Results on CIFAR-100 Dataset}
	
	We conduct experiments on CIFAR-100 dataset~\cite{krizhevsky2009learning} according to previous work~\cite{rao2023parameter}. The baseline methods include offline knowledge distillation methods (Vanilla~\cite{Hinton2015DistillingTK}, PKT~\cite{passalis2018learning}), representation knowledge distillation methods (CRD~\cite{tian2019contrastive}, RKD~\cite{park2019relational}), and online knowledge distillation methods (DML~\cite{zhang2018deep}, KDCL~\cite{guo2020online}, FT-KD~\cite{rao2023parameter}, PESF-KD~\cite{rao2023parameter}). In the experiment, the combination of student and teacher networks includes ResNet56-ResNet20, ResNet110-ResNet32, ResNet56-VGG8, and VGG13-VGG8. 
	
	We use the same training setup as described in \cite{rao2023parameter}, employing an SGD optimizer with a momentum of 0.9, a batch size of 64, a weight decay of $5 \times 10^{-4}$. The initial learning rate is set at 0.05, with a decay factor of 10 every 30 epochs starting from epoch 150. For the AMB module, we specify a learning rate of $1\times10^{-4}$. The comparison results with three different random seeds, presented as acc(@1, mean/standard deviation), are shown in Table~\ref{table3}. Our AMBOKD achieve the best performance among all compared methods, demonstrating its robustness and practicability.
	
	\begin{table}
		\centering
		\renewcommand{\arraystretch}{1.1}
		\caption{Comparison results on CIFAR-100 Test set}
		\setlength{\tabcolsep}{3pt}
		\resizebox{1\linewidth}{!}
		{
			
			\begin{tabular}{ccccc}
				\toprule
				\multirow{2}*{Method} & \multicolumn{4}{c}{ACC/Std (\%)}\cr
				\cmidrule(lr){2-5}
				& ResNet56-20 & ResNet110-32 & VGG13-8 & ResNet56-VGG8 \cr
				
				\midrule
				Vanilla~\cite{Hinton2015DistillingTK} & 70.95/0.51 & 73.08/0.42 & 73.36/0.24 & 73.98/0.33 \cr
				PKT~\cite{passalis2018learning} & 71.27/- & 73.67/- & 73.40/- & 74.10/- \cr
				CRD~\cite{tian2019contrastive} & 71.44/- & 73.62/- & 73.31/- & 74.06/- \cr
				RKD~\cite{park2019relational} & 71.47/- & 73.53/- & 74.15/- & 73.35/- \cr
				KDCL~\cite{guo2020online} & 70.11/- & 72.87/- & 73.99/- & 73.16/- \cr
				DML~\cite{zhang2018deep} & 71.40/- & 72.21/- & 74.18/- & 73.86/- \cr
				FT-KD~\cite{rao2023parameter} & 71.65/0.11 & 73.90/0.22 & 73.52/0.14 & 74.40/0.20 \cr
				PESF-KD~\cite{rao2023parameter} & 71.84/0.27 & 74.23/0.26 & 74.74/0.39 & 74.67/0.28 \cr
				AMBOKD (ours)& \textbf{72.26}/0.16 & \textbf{74.35}/0.27 & \textbf{74.95}/0.26 & \textbf{74.84}/0.27 \cr
				\bottomrule
			\end{tabular}
		}
		\label{table3}
	\end{table} 
	\subsubsection{Comparison Results on SEED-VIG Dataset}
	
	To evaluate the effectiveness and generalizability of our proposed method, we conducted experiments using the publicly available multimodal SEED-VIG dataset~\cite{zheng2017multimodal}. SEED-VIG is a comprehensive dataset designed for vigilance estimation in simulated driving scenarios. The dataset quantifies drivers’ psychological fatigue states using the PERCLOS annotation method, which ranges from 0 to 1, with a threshold of 0.35 used to distinguish between awake and fatigued states for binary classification. It includes 21 channels of EEG data and 4 channels of EOG data collected from 23 subjects, with each subject contributing 885 samples.
	
	In our experiments, we excluded data from subject 13 due to its imbalanced distribution and utilized preprocessed features from the SEED-VIG dataset as model inputs. The selected EEG features included Power Spectral Density (PSD) and Differential Entropy (DE), both processed with moving average filters across five frequency bands: delta (1$\sim$4 Hz), theta (4$\sim$8 Hz), alpha (8$\sim$14 Hz), beta (14$\sim$31 Hz), and gamma (31$\sim$50 Hz). The EOG features were extracted using three preprocessing methods: \texttt{'features table ICA'}, \texttt{'features table minus'}, and \texttt{'features table ICAv minh'}.
	
	The learning rate was set to 0.001, and the Adam optimizer with default parameter values was employed. The training process consisted of 30 epochs with a batch size of 64. A leave-one-out cross-validation approach was utilized for cross-subject experiments to evaluate the model’s generalization performance. EEGNet~\cite{lawhern2018eegnet} and EOGNet~\cite{cheng2022vigilancenet} were used as unimodal models and as feature extractors for the multimodal model. The comparative results, averaged over five different random seeds, are presented in Table~\ref{table5}. The results demonstrate that multimodal methods further improve performance metrics, with our proposed AMBOKD method achieving state-of-the-art performance, highlighting its effectiveness and generalizability in multimodal learning.
	
	\begin{table}
	\centering
	\renewcommand{\arraystretch}{1.1}
	\caption{Comparison results on SEED-VIG dataset}
	\setlength{\tabcolsep}{3pt}
	\resizebox{1\linewidth}{!}
	{
		
		\begin{tabular}{ccccc}
			\toprule
			Method & AUC/Std(\%) & F1/Std(\%) & ACC/Std(\%) & Precision/Std(\%)\cr
			\midrule
			EEGNet & 83.48/0.60 & 75.29/0.36 & 73.76/0.48 & 82.60/0.44 \cr
			EOGNet & 89.96/0.09 & 80.13/0.76 & 78.43/0.79 & 86.32/0.44 \cr
			\midrule
			MKD & 91.31/0.41 & 79.67/0.47 & 78.47/0.66 & 86.60/0.23 \cr
			EMKD & 91.36/0.44 & 79.70/0.46 & 78.51/0.64 & 86.62/0.21 \cr
			CA-MKD & 91.31/0.45 & 79.59/0.47 & 78.39/0.65 & 86.58/0.20 \cr
			C2KD & 90.94/0.18 & 81.88/0.24 & 80.37/0.30 & 86.95/0.17 \cr
			DIST+ & 91.26/0.35 & 83.36/0.73 & 81.78/0.82 & \textbf{88.08}/0.33 \cr
			DML\textsuperscript{*} & 90.15/0.26 & 81.37/0.57 & 79.96/0.61 & 86.55/0.23 \cr
			KDCL\textsuperscript{*} & 89.63/0.39 & 81.16/0.56 & 79.66/0.62 & 85.96/0.43\cr
			EML\textsuperscript{*} & 90.86/0.30 & 81.59/0.29 & 80.05/0.27 & 87.01/0.24\cr
			MLEMKD\textsuperscript{*} & 90.73/0.37 & 81.85/0.28 & 80.35/0.36 & 86.98/0.29 \cr
			AMBOKD (ours)\textsuperscript{*} & \textbf{91.64}/0.16 & \textbf{84.34}/0.53 & \textbf{83.32}/0.63 & 88.04/0.24 \cr
			\bottomrule
		\end{tabular}
	}
	\label{table5}

\end{table} 
\begin{figure*}
	\centerline{\includegraphics[width=1.8\columnwidth]{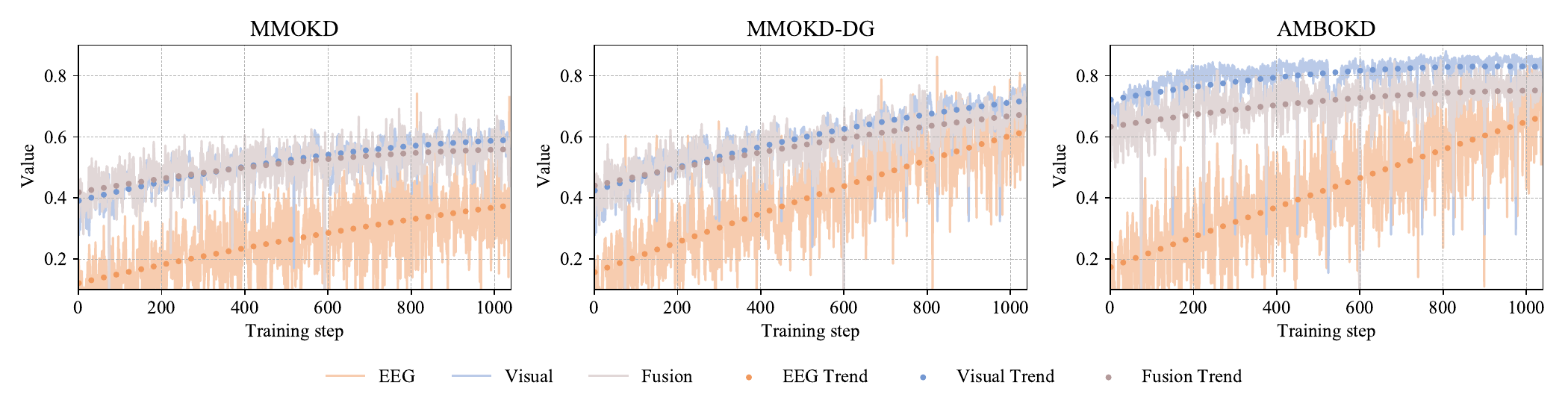}}
	\caption{Dynamic optimization degree changes of MMOKD, MMOKD-DG and AMBOKD during training process on the ESSVP dataset.}
	\label{figurelabel7}

\end{figure*}
	\subsection{Ablation Analysis}
	We conduct ablation experiments, performing data and visualization analyses on the results.
	\subsubsection{Data Analysis}
	The impact of incorporating OKD with mutual learning and AMB in multimodal learning is investigated through a detailed ablation analysis. Specifically, the role of each component is examined by progressively simplifying the proposed AMBOKD approach. MMOKD-DG and MMOKD-DK are the variants of our method, which keep the part of the dynamic weights block and the dynamic gradients block in the AMB module, separately. Then, MMOKD omits the entire AMB component, serving as a baseline to evaluate the underlying OKD and mutual learning mechanism. MKD further eliminates the online training method, representing a multi-teacher KD framework. The V-KD and E-KD approaches are further simplifications, using single-teacher KD from the visual and EEG modalities, respectively. Lastly, AMM is the most basic form of our model, which depends solely on a multi-head self-attention mechanism for modality fusion, without any KD. In addition, our analysis incorporates the unimodality encoders MCGRAM for EEG and EfﬁcientNet for the visual domain. 

	The ablation results are presented in Table~\ref{table1}. As can be seen, these uni-modality encoders MCGRAM and EfficientNet are outperformed by the AMM model, which leverages multi-head self-attention for fusion, thereby validating the multi-head self-attention mechanism's capacity to effectively integrate multimodal data. The results of E-KD, V-KD, and M-KD further verify the effectiveness of the knowledge distillation mechanism to improve the performance of the multimodal fusion model. MMOKD outperforms MKD by 1.37\% and 1.9\% in AUC, which demonstrates the effectiveness of the mutual learning method for achieving fully modal interaction and improving multimodal model performance. The MMOKD-DG and MMOKD-DK methods outperform the MMOKD method, indicating that adaptive gradient balancing and interactions between modality models provide a more effective training scheme. Our latest method, AMBOKD fuses the above two dynamic balancing schemes and achieves the optimal results, 2.08\% and 2.05\% higher than MMOKD, demonstrating the significant potential of modal balancing in multimodal mutual learning tasks.
	
	\begin{table}
		\centering
		\renewcommand{\arraystretch}{1.1}
		\caption{The ablation studies of different modules in subject-independent experiments}
		\label{tab3}
		\setlength{\tabcolsep}{3pt}
		\resizebox{1\linewidth}{!}
		{
			
			\begin{tabular}{cccccc}
				\toprule
				\multirow{2}*{Method} & \multicolumn{2}{c}{Session 1} & \multicolumn{2}{c}{Session 2}\cr
				\cmidrule(lr){2-3}
				\cmidrule(lr){4-5}
				& AUC/Std (\%) & \textit{p}-value (AUC) & AUC/Std (\%) & \textit{p}-value (AUC) \cr
				\midrule
				MCGRAM & 82.47/7.11 & $<10^{-3}$ & 80.44/5.32 & $<10^{-3}$ \cr
				EfficientNet & 78.81/8.89 & $<10^{-3}$ & 84.13/3.92 & $<10^{-3}$\cr
				AMM & 87.16/3.62 & $<10^{-3}$ & 87.75/3.64 & $<10^{-3}$ \cr
				EKD & 88.48/4.28 & $<10^{-3}$ & 88.68/3.73 & $<10^{-3}$ \cr
				VKD & 88.08/4.20 & $<10^{-3}$ & 87.53/4.76 & $<10^{-3}$ \cr
				MKD & 90.21/4.45 & $<10^{-3}$ & 89.35/3.54 & $<10^{-3}$ \cr
				MMOKD & 91.58/3.89 & $<10^{-3}$ & 91.25/2.47 & $<10^{-3}$ \cr
				MMOKD-DK & 91.90/3.48 & $<10^{-3}$ & 91.61/2.92 & $<10^{-3}$ \cr
				MMOKD-DG & 92.31/3.54 & $<10^{-3}$ & 92.59/2.21 & $<10^{-3}$ \cr
				AMBOKD (ours)& \textbf{93.66}/2.95 & $-$ & \textbf{93.30}/2.64 & $-$ \cr
				\bottomrule
			\end{tabular}
		}
		\label{table1}
		\vspace{-1em}
	\end{table}
	\subsubsection{Visualization Analysis}
	The latent representations of the models in the ablation analysis are visualized using the t-distributed stochastic neighbor embedding (t-SNE) approach to further verify the feature extraction and fusion capability of our proposed approach~\cite{liu2021using}. As depicted in Fig.~\ref{figurelabel6}, the feature distributions of all subjects are displayed by different colored dots, with red dots representing the features in the true target domain, and dots in other colors representing the features in the true non-target domain.
	
	As can be seen from Fig.~\ref{figurelabel6}, unimodality models such as EfficientNet and MCGRAM exhibit distinct visual and EEG modality characteristics in their feature representations. The visual-based EfficientNet model exhibits a more discrete feature representation, with a large intra-class distance, indicating its superior feature extraction performance but poor generalization performance of the classifier. Conversely, the feature representation of the MCGRAM model based on EEG is more clustered, with a small intra-class distance, indicating weaker feature extraction ability but stronger classifier generalization. As a multimodal fusion model, AMM effectively combines the benefits of vision and EEG modalities, thereby further reducing the inter-class distance while maintaining the feature representation capability. By integrating single-teacher guidance mechanisms into the multimodal fusion model, V-KD and E-KD models exhibit enhanced classification performance and generalization, as reflected in larger inter-class and smaller intra-class differences in feature representation. Furthermore, by introducing a multi-teacher mechanism, OKD, and the AMB module, the feature representations of MKD, MMOKD, and AMBOKD models exhibit enhanced discrimination and reduced intra-class distances. This indicates that these mechanisms enhance the capability of multimodal fusion and effectively improve the classifier's accuracy and generalization.
	
	In conclusion, the ablation study confirms the effectiveness of the AMBOKD framework and its modules. The proposed AMBOKD approach outperforms other methods in the multimodal fusion task, exhibiting robust discrimination between positive and negative samples, an increase in inter-class distance, a reduction in intra-class distance, and statistically significant experimental findings ($p<0.05$).

	\subsection{Adaptive Modality Balancing Analysis}
	
	In this section, we use the dynamic learning progress ratio $R^{DG}_{(t)}$(Eq.~\ref{eq14}), which is capable of showing the dynamic optimization level of each modality to assess the effectiveness of the AMB module. 
	
	As illustrated in Fig.~\ref{figurelabel7}, we draw the changes of $R^{DG}_{(t)}$ for each modality within MMOKD, MMOKD-DG, and AMBOKD during the training process on the ESSVP dataset, which can effectively indicate the real-time optimization degree of each modality. Although the standard MMOKD approach supports intermodal learning, its optimization is hindered by the inherent characteristics of each modality. This discrepancy in optimization results in a suboptimal mutual learning effect among the modalities, convergent feature vectors cannot be learned between modalities. The Dynamic Gradient (DG) block is able to address this issue by dynamically adjusting the training gradients according to each modality's learning progression. With this adjustment, MMOKD-DG improves the optimization speed and the final optimization degree of each modality. AMBOKD further employs a dynamic weight modulation module tailored to KL loss, guaranteeing the further optimization of higher performing modalities, while also supporting sustained development in the slower modality. This strategy ensures that all three modalities are optimally and uniformly enhanced throughout the training process, learning more consistent features and achieving superior overall performance.

	
	\vspace{-1em}
	\begin{table}
		\centering
		\renewcommand{\arraystretch}{1.1}
		\caption{The subject-independent results of the baseline and state-of-art methods in generalization analysis}
		\label{tab4}
		\setlength{\tabcolsep}{3pt}
		\resizebox{0.9\linewidth}{!}
		{
			
			\begin{tabular}{cccccc}
				\toprule
				\multirow{2}*{Method} & \multicolumn{2}{c}{Session 1} & \multicolumn{2}{c}{Session 2}\cr
				\cmidrule(lr){2-3}
				\cmidrule(lr){4-5}
				& AUC/Std (\%) & F1/Std (\%) & AUC/Std (\%) & F1/Std (\%) \cr
				\midrule
				MKD & 85.38/6.53 & 76.85/7.69 & 82.86/6.69 & 74.13/9.69 \cr
				EMKD & 86.56/6.65 & 77.08/7.49 & 83.61/7.24 & 74.78/9.04 \cr
				CA-MKD & 86.57/6.82 & 75.64/6.97 & 84.02/6.97 & 73.03/10.38 \cr
				C2KD & 78.68/8.64 & 64.20/8.24 & 74.93/7.39 & 59.22/11.05 \cr
				DIST+ & 86.53/6.28 & 71.55/7.28 & 82.85/6.88 & 66.26/6.84 \cr
				DML & 87.76/6.21 & 76.24/7.98 & 86.53/5.33 & 75.27/8.86 \cr
				KDCL & 88.13/6.44 & 77.75/7.32 & 86.91/5.29 & 74.98/7.92 \cr
				EML & 87.33/6.99 & 75.91/7.34 & 87.36/6.05 & 76.15/8.47 \cr
				MLEMKD & 84.40/5.70 & 71.94/6.16 & 84.23/4.48 & 72.47/6.80 \cr
				AMBOKD & \textbf{88.52}/5.57 & \textbf{80.34}/7.70 & \textbf{87.76}/5.53 & \textbf{78.98}/7.88 \cr
				\bottomrule
			\end{tabular}
		}

		\label{table2}
	\end{table} 
	
	\subsection{Generalization Analysis}
	\label{seq55}
	
	To further analyze the generalization performance of the AMBOKD method in different scenarios, we conduct transfer experiments from simple scenes to complex scenes under few-sample conditions. (the scene images are shown in Fig.~\ref{figurelabel5-1}). 
	
	In this experiment, we utilize the optimally saved model trained in the subject-independent experiment. This model is further refined using 30 real-world scene samples and is deployed to detect vehicle targets within these real scenes. The comparative results, presented in Table~\ref{table2}, demonstrate that the accuracy of our method decreases in complex scenes after transfer experiments, but it is still higher than all other comparison methods in both session 1 and session 2. These results confirm the effectiveness and generalization ability of our method.

	\begin{figure}
		\centerline{\includegraphics[width=1\columnwidth]{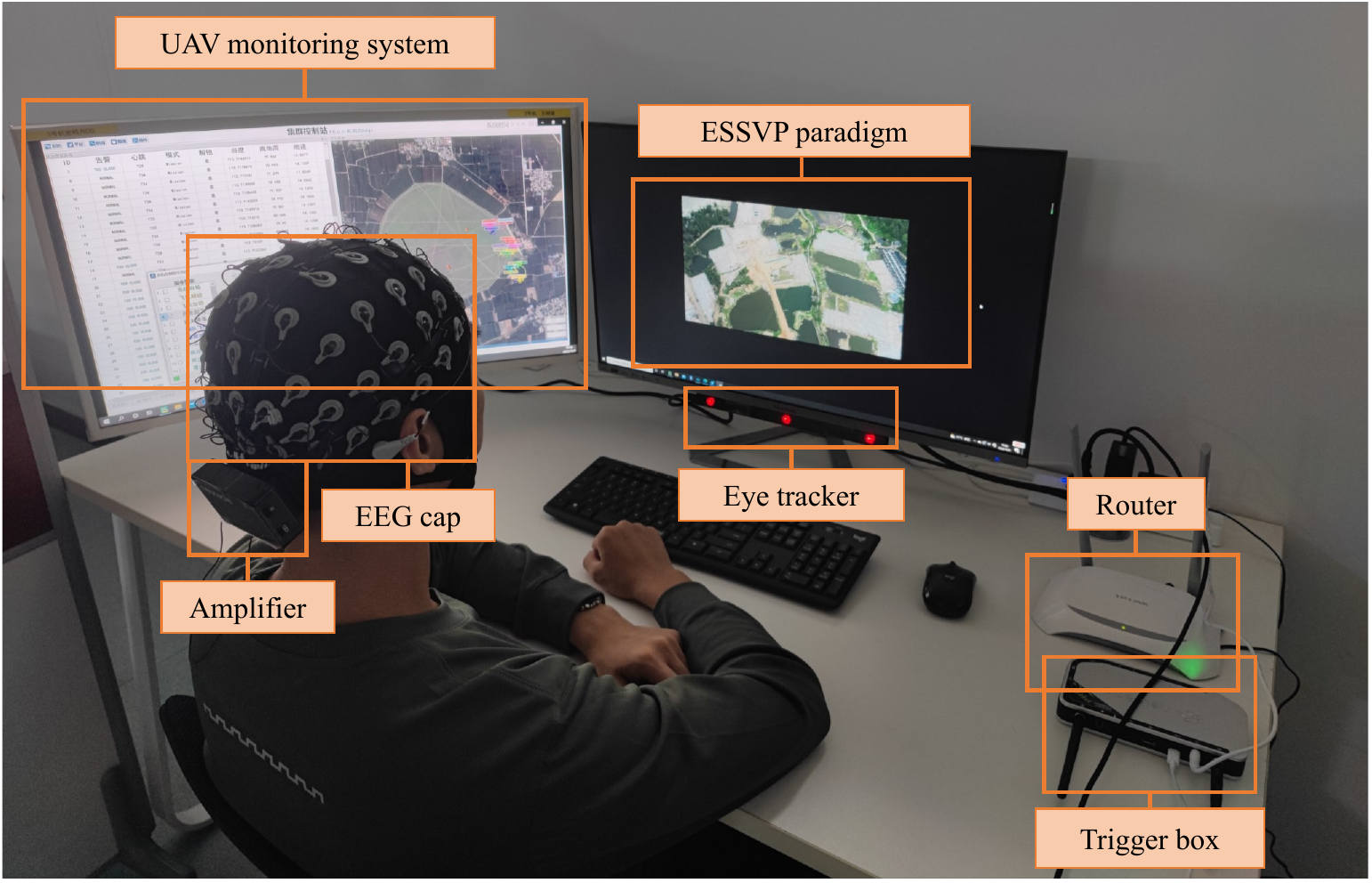}}
		\caption{The brain-eye-computer system verification for the task of dim object recognition while monitoring unmanned aerial vehicles.}
		\label{figurelabel10}
	\end{figure}
	\begin{figure}
		\centerline{\includegraphics[width=0.9\columnwidth]{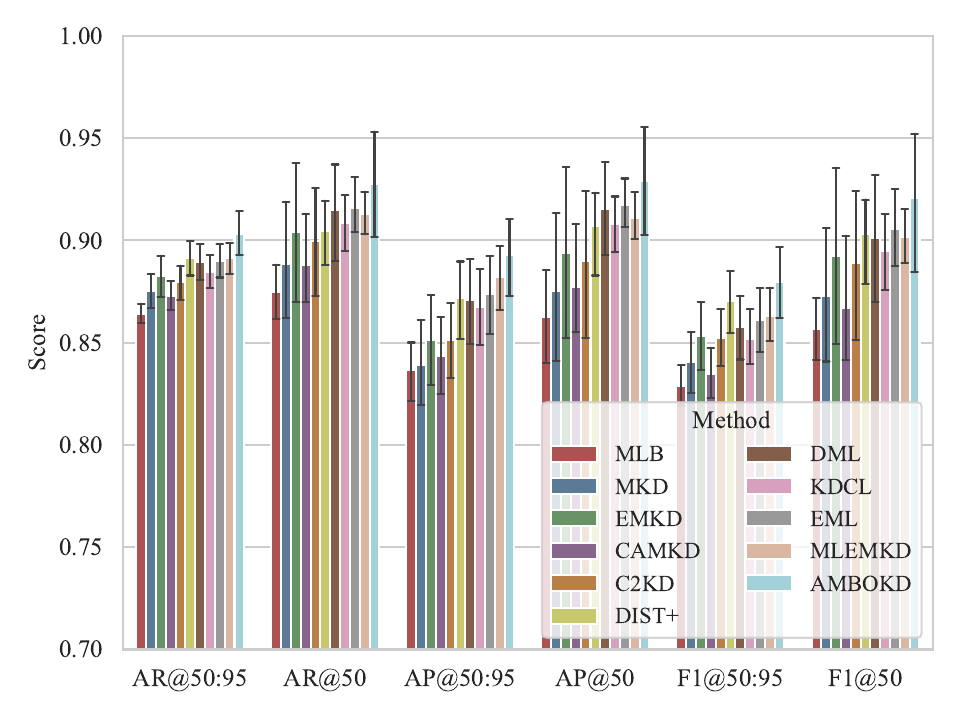}}
		\caption{The results on physical verification experiment.}
		\label{figurelabel11}
	\end{figure}
	\subsection{Physical Verification}

	To verify the practical effectiveness of our built system and the proposed method, we carry out our physical experiments combined with the application scenario of UAV ground control station operators, as illustrated in Fig.~\ref{figurelabel10}. During the system experiment, subjects are positioned in front of a computer screen with proper body alignment, focusing their gaze directly on the frontal display while performing the ESSVP paradigm. They are tasked with controlling the system and issuing appropriate commands to the UAV upon detecting abnormal displays on the UAV monitoring interface located to the participant's left. Concurrently, the ESSVP paradigm temporarily halts the playback of visual stimuli to accommodate this interaction. 

	In the system verification experiment, the models refined using 30 real-world scene samples are employed directly to process the multimodal data and detect the target. As shown in Fig.~\ref{figurelabel11}, the AMBOKD method achieves the highest performances in all metrics compared with the baseline and state-of-the-art methods. These experimental findings demonstrate the effectiveness of the brain-eye-computer based object detection system in real-world settings, indicating robustness and high accuracy in practical tasks.
		
	We further compared our brain-eye-computer based system with traditional manual and computer vision approaches, as shown in Table~\ref{table4}. In the human-based experiment, five subjects with no prior expertise in aerial image analysis were recruited to detect and annotate vehicle targets in 60 aerial images, ensuring a fair baseline for manual detection. Each subject underwent a 10-minute training session to familiarize themselves with the annotation guidelines, and the experiment was conducted in a controlled environment with consistent lighting and display settings to ensure uniformity. The results indicate significant variability in detection time and accuracy among subjects, with a mean accuracy of 78.20±18.20\% and a detection time of 1026.00±360.00s (as shown in Table VI). This high variance highlights the inconsistency of human performance in such tasks, underscoring the inefficiency and unreliability of purely manual methods in complex scenarios, as stable performance is challenging without automated assistance. In the computer-based experiment, we adopted the same training setup as the brain-eye-computer based system: a Faster RCNN model with a ResNet50 backbone, trained on the public dataset, was fine-tuned with 30 samples and then tested on 60 images. Although this method achieves rapid detection capabilities, its accuracy was notably compromised due to the limited training samples. In contrast, our integrated brain-eye-computer system achieved superior performance, registering a total processing time of 245.37s (240s for the ESSVP paradigm and 5.37s for data analysis). This result not only demonstrates the efficiency of the system but also its robust generalization capabilities across different test conditions.
	
	\begin{table}[t]
		\centering
		\renewcommand{\arraystretch}{1.1}
		\caption{Comparison between human, computer, and the brain-eye-computer based system}
		\label{tab5}
		\setlength{\tabcolsep}{3pt}
		\resizebox{1\linewidth}{!}
		{
			
			\begin{tabular}{cccccc}
				\toprule
				Method & AP@50:95 (\%) & AR@50:95 (\%) & F1@50:95 (\%) & Total Time (s)\cr
				\midrule
				Human & 78.28$\pm$18.20 & 67.33$\pm$20.26 & 72.00$\pm$18.71 & 1026.00$\pm$362.00\cr
				Computer  & 23.44$\pm$5.49 & 30.12$\pm$6.89 & 26.36$\pm$6.10 & \textbf{3.00}$\pm$0 \cr
				Brain-eye-computer & \textbf{89.26}$\pm$6.76 & \textbf{90.32}$\pm$4.01 & \textbf{87.98}$\pm$6.37 & 245.37$\pm$1.38 \cr
				\bottomrule
			\end{tabular}
		}
		\label{table4}
	\end{table} 
	\section{Conclusion and Future Work}
	\label{seq6}	
	This study builds a brain-eye-computer based system for object detection in aerial images under few-shot conditions. This system uses a region proposal network to detect suspicious targets in aerial images. After obtaining the images with the region proposals, it elicits subjects' ERP signals during target search with the ESSVP paradigm and constructs EEG-image data pairs incorporating eye movement data. These pairs are then recognized using the proposed AMBOKD method. AMBOKD fully extracts and fuses crucial information from EEG and visual modality features, facilitates end-to-end mutual learning, and improves adaptive multimodal interaction capability through the AMB module. Experimental results demonstrate the effectiveness and superiority of our method, as well as its ability to enhance unimodal model performance. Lastly, the feasibility and transferability of the AMBOKD method and the brain-eye-computer based system are verified through experiments with practical scenario images. This study opens new possibilities for robust and efficient dim object detection in aerial applications.
	
	Despite its promising results, the proposed method has several limitations that should be addressed in future research. First, the current system is designed and evaluated in offline settings, which may limit its real-time applicability in fast-paced aerial detection tasks. Future efforts will focus on developing real-time online systems to improve response time and practical utility across a broader range of scenarios. Second, the system’s performance may vary based on the subject’s familiarity and engagement with the ESSVP paradigm, necessitating more adaptive and efficient experimental paradigms to reduce inter-subject variability. Additionally, to further reduce experimental and application costs, future work will explore system experiments based on low-cost EEG devices with fewer channels. Finally, while the study recruited a controlled sample to ensure consistency, the homogeneity of participants may affect generalizability. Expanding the dataset with more diverse subjects and aerial images will help enhance robustness and applicability in real-world scenarios.
	
	\bibliographystyle{IEEEtran}
	\bibliography{IEEEabrv, refs.bib}

\begin{thebibliography}{10}
\providecommand{\url}[1]{#1}
\csname url@samestyle\endcsname
\providecommand{\newblock}{\relax}
\providecommand{\bibinfo}[2]{#2}
\providecommand{\BIBentrySTDinterwordspacing}{\spaceskip=0pt\relax}
\providecommand{\BIBentryALTinterwordstretchfactor}{4}
\providecommand{\BIBentryALTinterwordspacing}{\spaceskip=\fontdimen2\font plus
\BIBentryALTinterwordstretchfactor\fontdimen3\font minus
  \fontdimen4\font\relax}
\providecommand{\BIBforeignlanguage}[2]{{%
\expandafter\ifx\csname l@#1\endcsname\relax
\typeout{** WARNING: IEEEtran.bst: No hyphenation pattern has been}%
\typeout{** loaded for the language `#1'. Using the pattern for}%
\typeout{** the default language instead.}%
\else
\language=\csname l@#1\endcsname
\fi
#2}}
\providecommand{\BIBdecl}{\relax}
\BIBdecl

\bibitem{9451544}
Z.~Li, F.~Liu, W.~Yang, S.~Peng, and J.~Zhou, ``A survey of convolutional
  neural networks: Analysis, applications, and prospects,'' \emph{IEEE
  Transactions on Neural Networks and Learning Systems}, vol.~33, no.~12, pp.
  6999--7019, 2022.

\bibitem{10.1109/TPAMI.2023.3290594}
G.~Cheng, X.~Yuan, X.~Yao, K.~Yan, Q.~Zeng, X.~Xie, and J.~Han, ``Towards
  large-scale small object detection: Survey and benchmarks,'' \emph{IEEE
  Transactions on Pattern Analysis and Machine Intelligence}, vol.~45, no.~11,
  p. 13467–13488, 2023.

\bibitem{jia2022low}
Z.~Jia, X.~Xu, J.~Hu, and Y.~Shi, ``Low-power object-detection challenge on
  unmanned aerial vehicles,'' \emph{Nature Machine Intelligence}, vol.~4,
  no.~12, pp. 1265--1266, 2022.

\bibitem{lan2021macro}
Z.~Lan, C.~Yan, Z.~Li, D.~Tang, and X.~Xiang, ``{MACRO}: multi-attention
  convolutional recurrent model for subject-independent {ERP} detection,''
  \emph{IEEE Signal Processing Letters}, vol.~28, pp. 1505--1509, 2021.

\bibitem{li2022mcgram}
Z.~Li, C.~Yan, Z.~Lan, D.~Tang, and X.~Xiang, ``{MCGRAM}: Linking multi-scale
  cnn with a graph-based recurrent attention model for subject-independent
  {ERP} detection,'' \emph{IEEE Transactions on Circuits and Systems II:
  Express Briefs}, vol.~69, no.~12, pp. 5199--5203, 2022.

\bibitem{Lees2018ARO}
S.~Lees, N.~Dayan, H.~Cecotti, P.~Mccullagh, L.~P. Maguire, F.~Lotte, and D.~H.
  Coyle, ``A review of rapid serial visual presentation-based brain–computer
  interfaces,'' \emph{Journal of Neural Engineering}, vol.~15, 2018.

\bibitem{quan2024multimodal}
S.~Quan, J.~Yan, K.~Guo, Y.~Zheng, M.~Dong, and J.~Liang, ``Multimodal
  contrastive learning for brain--machine fusion: From brain-in-the-loop
  modeling to brain-out-of-the-loop application,'' \emph{Information Fusion},
  vol. 110, p. 102447, 2024.

\bibitem{wang2021native}
Y.~Wang, J.~Yan, Z.~Yin, S.~Ren, M.~Dong, C.~Zheng, W.~Zhang, and J.~Liang,
  ``How native background affects human performance in real-world visual object
  detection: an event-related potential study,'' \emph{Frontiers in
  neuroscience}, vol.~15, p. 665084, 2021.

\bibitem{baltruvsaitis2018multimodal}
T.~Baltru{\v{s}}aitis, C.~Ahuja, and L.-P. Morency, ``Multimodal machine
  learning: A survey and taxonomy,'' \emph{IEEE Transactions on Pattern
  Analysis and Machine Intelligence}, vol.~41, no.~2, pp. 423--443, 2018.

\bibitem{10024142}
L.~Chen, M.~Li, M.~Wu, W.~Pedrycz, and K.~Hirota, ``Coupled multimodal
  emotional feature analysis based on broad-deep fusion networks in
  human–robot interaction,'' \emph{IEEE Transactions on Neural Networks and
  Learning Systems}, vol.~35, no.~7, pp. 9663--9673, 2024.

\bibitem{nagrani2021attention}
A.~Nagrani, S.~Yang, A.~Arnab, A.~Jansen, C.~Schmid, and C.~Sun, ``Attention
  bottlenecks for multimodal fusion,'' \emph{Advances in Neural Information
  Processing Systems}, vol.~34, pp. 14\,200--14\,213, 2021.

\bibitem{yang2022multimodal}
R.~Yang, S.~Wang, Y.~Sun, H.~Zhang, Y.~Liao, Y.~Gu, B.~Hou, and L.~Jiao,
  ``Multimodal fusion remote sensing image--audio retrieval,'' \emph{IEEE
  Journal of Selected Topics in Applied Earth Observations and Remote Sensing},
  vol.~15, pp. 6220--6235, 2022.

\bibitem{Hinton2015DistillingTK}
G.~Hinton, O.~Vinyals, and J.~Dean, ``Distilling the knowledge in a neural
  network,'' in \emph{NIPS Deep Learning and Representation Learning Workshop},
  2015.

\bibitem{Zhang2022VisualtoEEGCK}
S.~Zhang, C.~Tang, and C.~Guan, ``Visual-to-{EEG} cross-modal knowledge
  distillation for continuous emotion recognition,'' \emph{Pattern
  Recognition}, vol. 130, p. 108833, 2022.

\bibitem{zhang2022confidence}
H.~Zhang, D.~Chen, and C.~Wang, ``Confidence-aware multi-teacher knowledge
  distillation,'' in \emph{IEEE International Conference on Acoustics, Speech
  and Signal Processing (ICASSP)}.\hskip 1em plus 0.5em minus 0.4em\relax IEEE,
  2022, pp. 4498--4502.

\bibitem{he2016deep}
K.~He, X.~Zhang, S.~Ren, and J.~Sun, ``Deep residual learning for image
  recognition,'' in \emph{Proceedings of the IEEE/CVF Conference on Computer
  Vision and Pattern Recognition (CVPR)}, 2016, pp. 770--778.

\bibitem{sandler2018mobilenetv2}
M.~Sandler, A.~Howard, M.~Zhu, A.~Zhmoginov, and L.~Chen, ``Mobilenetv2:
  Inverted residuals and linear bottlenecks,'' in \emph{Proceedings of the
  IEEE/CVF Conference on Computer Vision and Pattern Recognition (CVPR)}, 2018,
  pp. 4510--4520.

\bibitem{tan2019efficientnet}
M.~Tan and Q.~Le, ``Efficientnet: Rethinking model scaling for convolutional
  neural networks,'' in \emph{International Conference on Machine Learning},
  2019, pp. 6105--6114.

\bibitem{zheng2021attention}
X.~Zheng and W.~Chen, ``An attention-based bi-{LSTM} method for visual object
  classification via {EEG},'' \emph{Biomedical Signal Processing and Control},
  vol.~63, p. 102174, 2021.

\bibitem{fan2022dc}
L.~Fan, H.~Shen, F.~Xie, J.~Su, Y.~Yu, and D.~Hu, ``Dc-tcnn: A deep model for
  {EEG}-based detection of dim targets,'' \emph{IEEE Transactions on Neural
  Systems and Rehabilitation Engineering}, vol.~30, pp. 1727--1736, 2022.

\bibitem{zhang2020benchmark}
S.~Zhang, Y.~Wang, L.~Zhang, and X.~Gao, ``A benchmark dataset for {RSVP}-based
  brain--computer interfaces,'' \emph{Frontiers in Neuroscience}, vol.~14, p.
  568000, 2020.

\bibitem{barngrover2015brain}
C.~Barngrover, A.~Althoff, P.~DeGuzman, and R.~Kastner, ``A brain--computer
  interface ({BCI}) for the detection of mine-like objects in sidescan sonar
  imagery,'' \emph{IEEE Journal of Oceanic Engineering}, vol.~41, no.~1, pp.
  123--138, 2015.

\bibitem{song2024decoding}
Y.~Song, B.~Liu, X.~Li, N.~Shi, Y.~Wang, and X.~Gao, ``Decoding {{Natural
  Images}} from {{EEG}} for {{Object Recognition}},'' in \emph{International
  {{Conference}} on {{Learning Representations}}}, 2024.

\bibitem{schrimpf2020integrative}
M.~Schrimpf, J.~Kubilius, M.~J. Lee, N.~A.~R. Murty, R.~Ajemian, and J.~J.
  DiCarlo, ``Integrative benchmarking to advance neurally mechanistic models of
  human intelligence,'' \emph{Neuron}, vol. 108, no.~3, pp. 413--423, 2020.

\bibitem{10089190}
C.~Du, K.~Fu, J.~Li, and H.~He, ``Decoding visual neural representations by
  multimodal learning of brain-visual-linguistic features,'' \emph{IEEE
  Transactions on Pattern Analysis and Machine Intelligence}, vol.~45, no.~9,
  pp. 10\,760--10\,777, 2023.

\bibitem{zhou2022mtanet}
W.~Zhou, S.~Dong, J.~Lei, and L.~Yu, ``{MTANet}: Multitask-aware network with
  hierarchical multimodal fusion for {RGB-T} urban scene understanding,''
  \emph{IEEE Transactions on Intelligent Vehicles}, vol.~8, no.~1, pp. 48--58,
  2022.

\bibitem{10123038}
P.~Xu, X.~Zhu, and D.~A. Clifton, ``Multimodal learning with transformers: A
  survey,'' \emph{IEEE Transactions on Pattern Analysis and Machine
  Intelligence}, vol.~45, no.~10, pp. 12\,113--12\,132, 2023.

\bibitem{chen2022dearkd}
X.~Chen, Q.~Cao, Y.~Zhong, J.~Zhang, S.~Gao, and D.~Tao, ``Dearkd:
  data-efficient early knowledge distillation for vision transformers,'' in
  \emph{Proceedings of the IEEE/CVF Conference on Computer Vision and Pattern
  Recognition (CVPR)}, 2022, pp. 12\,052--12\,062.

\bibitem{ryu2022knowledge}
M.~Ryu, G.~Lee, and K.~Lee, ``Knowledge distillation for bert unsupervised
  domain adaptation,'' \emph{Knowledge and Information Systems}, vol.~64,
  no.~11, pp. 3113--3128, 2022.

\bibitem{li2023decoupled}
Y.~Li, Y.~Wang, and Z.~Cui, ``Decoupled multimodal distilling for emotion
  recognition,'' in \emph{Proceedings of the IEEE/CVF Conference on Computer
  Vision and Pattern Recognition (CVPR)}, 2023, pp. 6631--6640.

\bibitem{gou2021knowledge}
J.~Gou, B.~Yu, S.~J. Maybank, and D.~Tao, ``Knowledge distillation: A survey,''
  \emph{International Journal of Computer Vision}, vol. 129, pp. 1789--1819,
  2021.

\bibitem{zhang2018deep}
Y.~Zhang, T.~Xiang, T.~M. Hospedales, and H.~Lu, ``Deep mutual learning,'' in
  \emph{Proceedings of the IEEE/CVF Conference on Computer Vision and Pattern
  Recognition (CVPR)}, 2018, pp. 4320--4328.

\bibitem{you2017learning}
S.~You, C.~Xu, C.~Xu, and D.~Tao, ``Learning from multiple teacher networks,''
  in \emph{Proceedings of the 23rd ACM SIGKDD International Conference on
  Knowledge Discovery and Data Mining}, 2017, pp. 1285--1294.

\bibitem{kwon2020adaptive}
K.~Kwon, H.~Na, H.~Lee, and N.~S. Kim, ``Adaptive knowledge distillation based
  on entropy,'' in \emph{IEEE International Conference on Acoustics, Speech and
  Signal Processing (ICASSP)}, 2020, pp. 7409--7413.

\bibitem{li2023embedded}
C.~Li, G.~Li, H.~Zhang, and D.~Ji, ``Embedded mutual learning: A novel online
  distillation method integrating diverse knowledge sources,'' \emph{Applied
  Intelligence}, vol.~53, no.~10, pp. 11\,524--11\,537, 2023.

\bibitem{9926046}
J.~Gou, L.~Sun, B.~Yu, L.~Du, K.~Ramamohanarao, and D.~Tao, ``Collaborative
  knowledge distillation via multiknowledge transfer,'' \emph{IEEE Transactions
  on Neural Networks and Learning Systems}, vol.~35, no.~5, pp. 6718--6730,
  2024.

\bibitem{ren2015faster}
S.~Ren, K.~He, R.~Girshick, and J.~Sun, ``Faster r-cnn: Towards real-time
  object detection with region proposal networks,'' \emph{IEEE Transactions on
  Pattern Analysis and Machine Intelligence}, vol.~39, no.~6, pp. 1137--1149,
  2017.

\bibitem{cecotti2011impact}
H.~Cecotti, J.~Sato-Reinhold, J.~L. Sy, J.~C. Elliott, M.~P. Eckstein, and
  B.~Giesbrecht, ``Impact of target probability on single-trial eeg target
  detection in a difficult rapid serial visual presentation task,'' in
  \emph{2011 Annual International Conference of the IEEE Engineering in
  Medicine and Biology Society}.\hskip 1em plus 0.5em minus 0.4em\relax IEEE,
  2011, pp. 6381--6384.

\bibitem{chatrian1985ten}
G.~E. Chatrian, E.~Lettich, and P.~L. Nelson, ``Ten percent electrode system
  for topographic studies of spontaneous and evoked {EEG} activities,''
  \emph{American Journal of EEG Technology}, vol.~25, no.~2, pp. 83--92, 1985.

\bibitem{vaswani2017attention}
A.~Vaswani, N.~Shazeer, N.~Parmar, J.~Uszkoreit, L.~Jones, A.~N. Gomez,
  {\L}.~Kaiser, and I.~Polosukhin, ``Attention is all you need,'' in
  \emph{Proceedings of the 31st International Conference on Neural Information
  Processing Systems}, 2017, p. 6000–6010.

\bibitem{peng2022balanced}
X.~Peng, Y.~Wei, A.~Deng, D.~Wang, and D.~Hu, ``Balanced multimodal learning
  via on-the-fly gradient modulation,'' in \emph{Proceedings of the IEEE/CVF
  Conference on Computer Vision and Pattern Recognition (CVPR)}, 2022, pp.
  8238--8247.

\bibitem{yang2023online}
C.~Yang, Z.~An, H.~Zhou, F.~Zhuang, Y.~Xu, and Q.~Zhang, ``Online knowledge
  distillation via mutual contrastive learning for visual recognition,''
  \emph{IEEE Transactions on Pattern Analysis and Machine Intelligence}, pp.
  10\,212--10\,227, 2023.

\bibitem{lawhern2018eegnet}
V.~J. Lawhern, A.~J. Solon, N.~R. Waytowich, S.~M. Gordon, C.~P. Hung, and
  B.~J. Lance, ``{EEGNet}: a compact convolutional neural network for
  {EEG}-based brain--computer interfaces,'' \emph{Journal of Neural
  Engineering}, vol.~15, no.~5, p. 056013, 2018.

\bibitem{huo2024c2kd}
F.~Huo, W.~Xu, J.~Guo, H.~Wang, and S.~Guo, ``C2kd: Bridging the modality gap
  for cross-modal knowledge distillation,'' in \emph{Proceedings of the
  IEEE/CVF Conference on Computer Vision and Pattern Recognition}, 2024, pp.
  16\,006--16\,015.

\bibitem{huang2025dist+}
T.~Huang, S.~You, F.~Wang, C.~Qian, and C.~Xu, ``Dist+: Knowledge distillation
  from a stronger adaptive teacher,'' \emph{IEEE Transactions on Pattern
  Analysis and Machine Intelligence}, 2025.

\bibitem{guo2020online}
Q.~Guo, X.~Wang, Y.~Wu, Z.~Yu, D.~Liang, X.~Hu, and P.~Luo, ``Online knowledge
  distillation via collaborative learning,'' in \emph{Proceedings of the
  IEEE/CVF Conference on Computer Vision and Pattern Recognition (CVPR)}, 2020,
  pp. 11\,020--11\,029.

\bibitem{10852525}
Y.~Qian, X.~Wang, F.~Sun, and L.~Pan, ``Compressing transfer: Mutual
  learning-empowered knowledge distillation for temporal knowledge graph
  reasoning,'' \emph{IEEE Transactions on Neural Networks and Learning
  Systems}, pp. 1--13, 2025.

\bibitem{krizhevsky2009learning}
A.~Krizhevsky, G.~Hinton \emph{et~al.}, ``Learning multiple layers of features
  from tiny images,'' \emph{University of Toronto}, 05 2012.

\bibitem{rao2023parameter}
J.~Rao, X.~Meng, L.~Ding, S.~Qi, X.~Liu, M.~Zhang, and D.~Tao,
  ``Parameter-efficient and student-friendly knowledge distillation,''
  \emph{IEEE Transactions on Multimedia}, 2023.

\bibitem{passalis2018learning}
N.~Passalis and A.~Tefas, ``Learning deep representations with probabilistic
  knowledge transfer,'' in \emph{Proceedings of the European Conference on
  Computer Vision (ECCV)}, 2018, pp. 268--284.

\bibitem{tian2019contrastive}
Y.~Tian, D.~Krishnan, and P.~Isola, ``Contrastive representation
  distillation,'' in \emph{arXiv preprint arXiv:1910.10699}, 2019.

\bibitem{park2019relational}
W.~Park, D.~Kim, Y.~Lu, and M.~Cho, ``Relational knowledge distillation,'' in
  \emph{Proceedings of the IEEE/CVF Conference on Computer Vision and Pattern
  Recognition (CVPR)}, 2019, pp. 3967--3976.

\bibitem{zheng2017multimodal}
W.-L. Zheng and B.-L. Lu, ``A multimodal approach to estimating vigilance using
  eeg and forehead eog,'' \emph{Journal of neural engineering}, vol.~14, no.~2,
  p. 026017, 2017.

\bibitem{cheng2022vigilancenet}
X.~Cheng, W.~Wei, C.~Du, S.~Qiu, S.~Tian, X.~Ma, and H.~He, ``Vigilancenet:
  Decouple intra-and inter-modality learning for multimodal vigilance
  estimation in rsvp-based bci,'' in \emph{Proceedings of the 30th ACM
  International Conference on Multimedia}, 2022, pp. 209--217.

\bibitem{liu2021using}
H.~Liu, J.~Yang, M.~Ye, S.~C. James, Z.~Tang, J.~Dong, and T.~Xing, ``Using
  t-distributed stochastic neighbor embedding (t-{SNE}) for cluster analysis
  and spatial zone delineation of groundwater geochemistry data,''
  \emph{Journal of Hydrology}, vol. 597, p. 126146, 2021.

\end{thebibliography}
	\vspace{-4em}
	\begin{IEEEbiography}[{\includegraphics[width=1in,height=1.25in, clip,keepaspectratio]{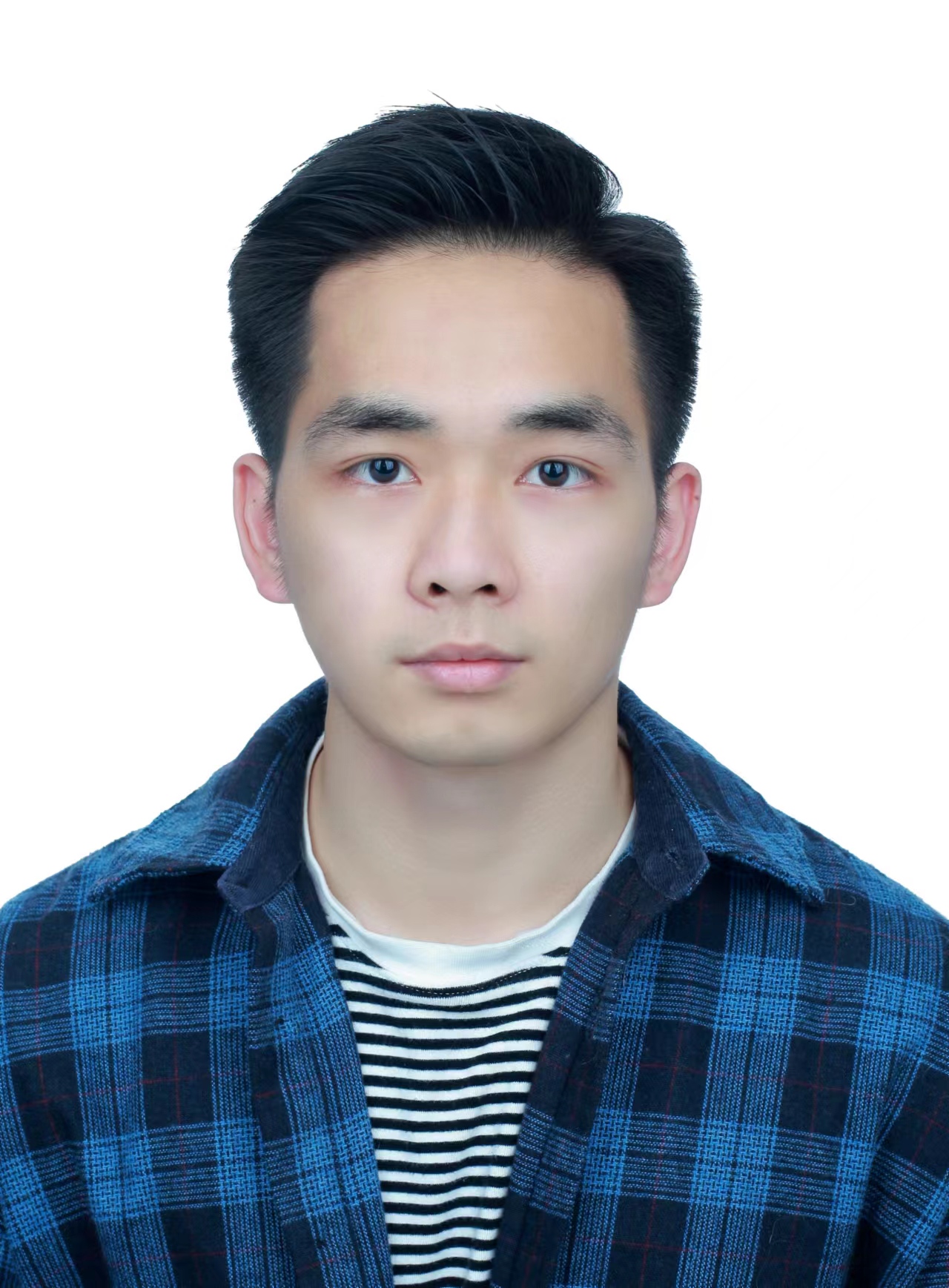}}]{Zixing Li} received the B.E. degree in electronics engineering and M.S degree in electronic information from the National University of Defense Technology, Changsha, China, in 2020 and 2022. He is currently pursuing the Ph.D. degree in control science and engineering with the College of Intelligence Science and Technology at the National University of Defense Technology.
		His research interests include brain-computer interface, EEG signal processing, and multi-modal learning.
		
	\end{IEEEbiography}
	
	\begin{IEEEbiography}[{\includegraphics[width=1in,height=1.25in, clip, keepaspectratio]{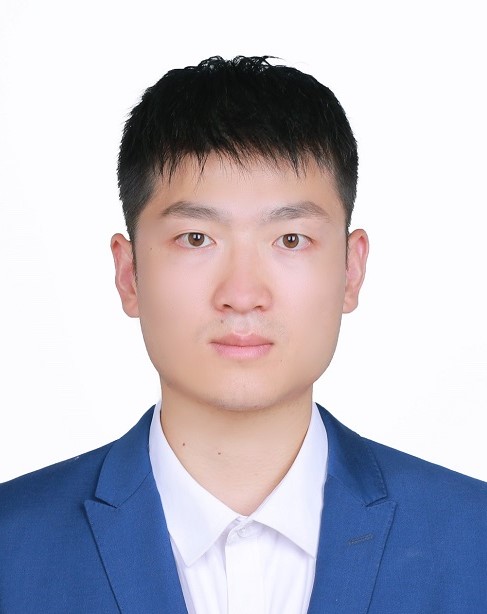}}]{Chao Yan} received the B.E. degree in electrical engineering and automation from China University of Mining and Technology, Xuzhou, China, in 2017, and the M.S. and Ph.D. degrees in control science and engineering from the National University of Defense Technology, Changsha, China, in 2019, and 2023, respectively. He was a visiting Ph.D. student with the School of Mechanical and Aerospace Engineering, Nanyang Technological University, Singapore, from 2021 to 2022. He is currently an Associate Professor with the College of Automation Engineering, Nanjing University of Aeronautics and Astronautics, Nanjing, China.
		His research interests include deep reinforcement learning, coordination control of UAV swarms.
	\end{IEEEbiography}
	
	\begin{IEEEbiography}[{\includegraphics[width=1in,height=1.25in, clip,keepaspectratio]{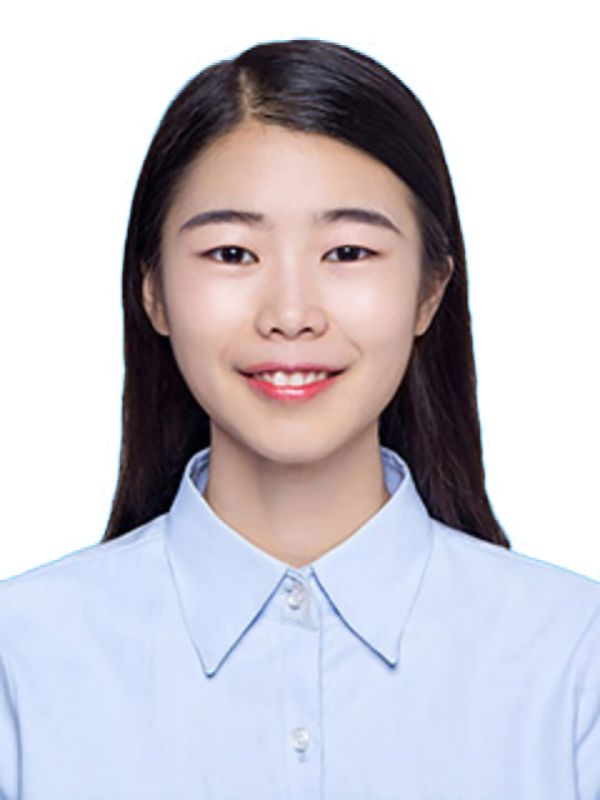}}]{Zhen Lan} received the B.E. degree in automation from the School of Automation Engineering, University of Electronic Science and Technology of China, Chengdu, China, in 2019. She is currently pursuing the Ph.D. degree in control science and engineering with the College of Intelligence Science and Technology, National University of Defense Technology, Changsha, China.
		Her research interests include brain-computer interface, multi-modal learning, and UAV control.
	\end{IEEEbiography}
	
	\begin{IEEEbiography}[{\includegraphics[width=1in,height=1.25in, clip,keepaspectratio]{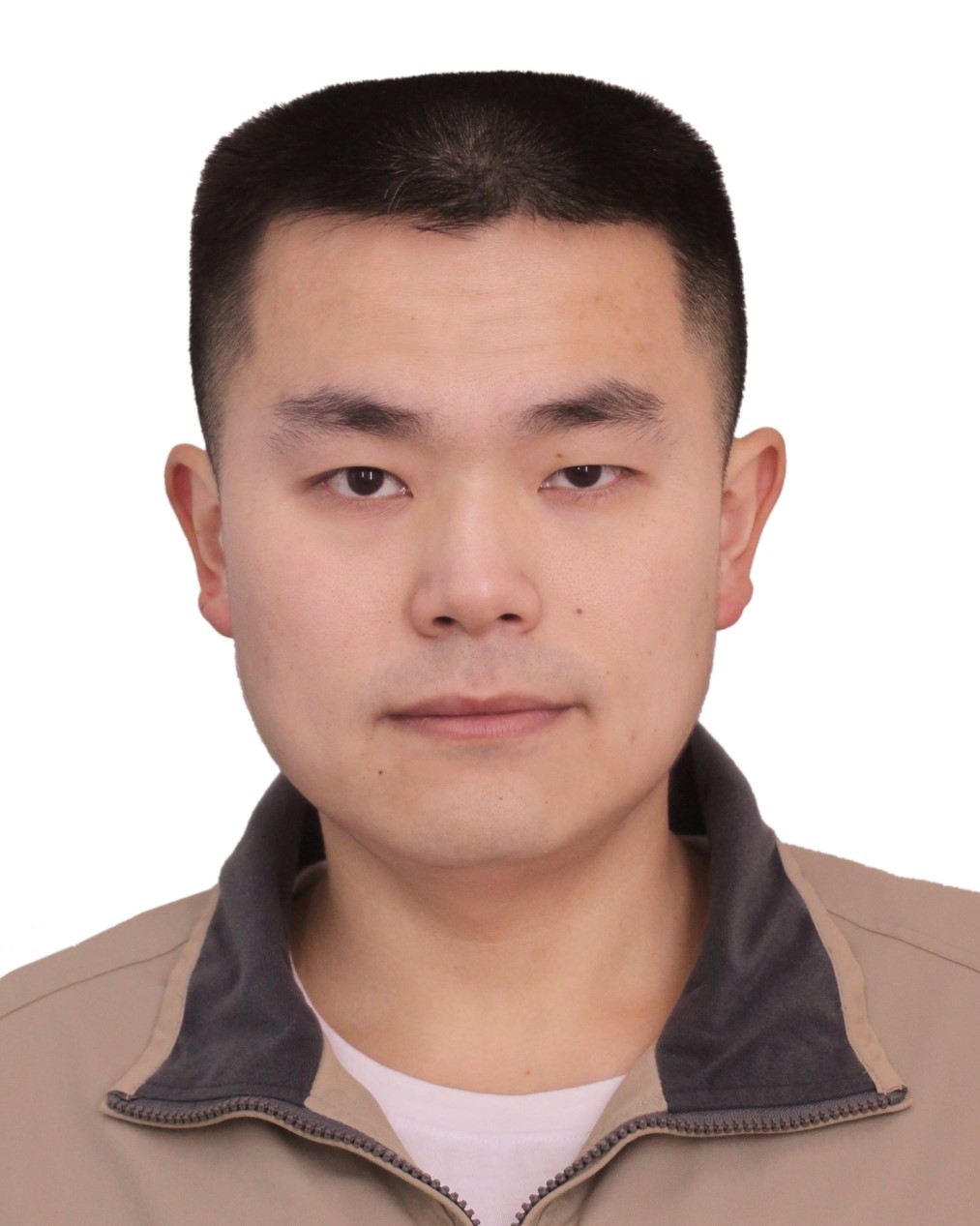}}]{Xiaojia Xiang} received the B.E., M.S., and Ph.D. degrees in control science and engineering from the National University of Defense Technology (NUDT), Changsha, China, in 2003, 2007, and 2016, respectively. 
		He is currently a Professor at the College of Intelligence Science and Technology, NUDT. His research interests include mission planning, autonomous and cooperative control of unmanned systems.
	\end{IEEEbiography}
	
	\begin{IEEEbiography}[{\includegraphics[width=1in,height=1.25in, clip,keepaspectratio]{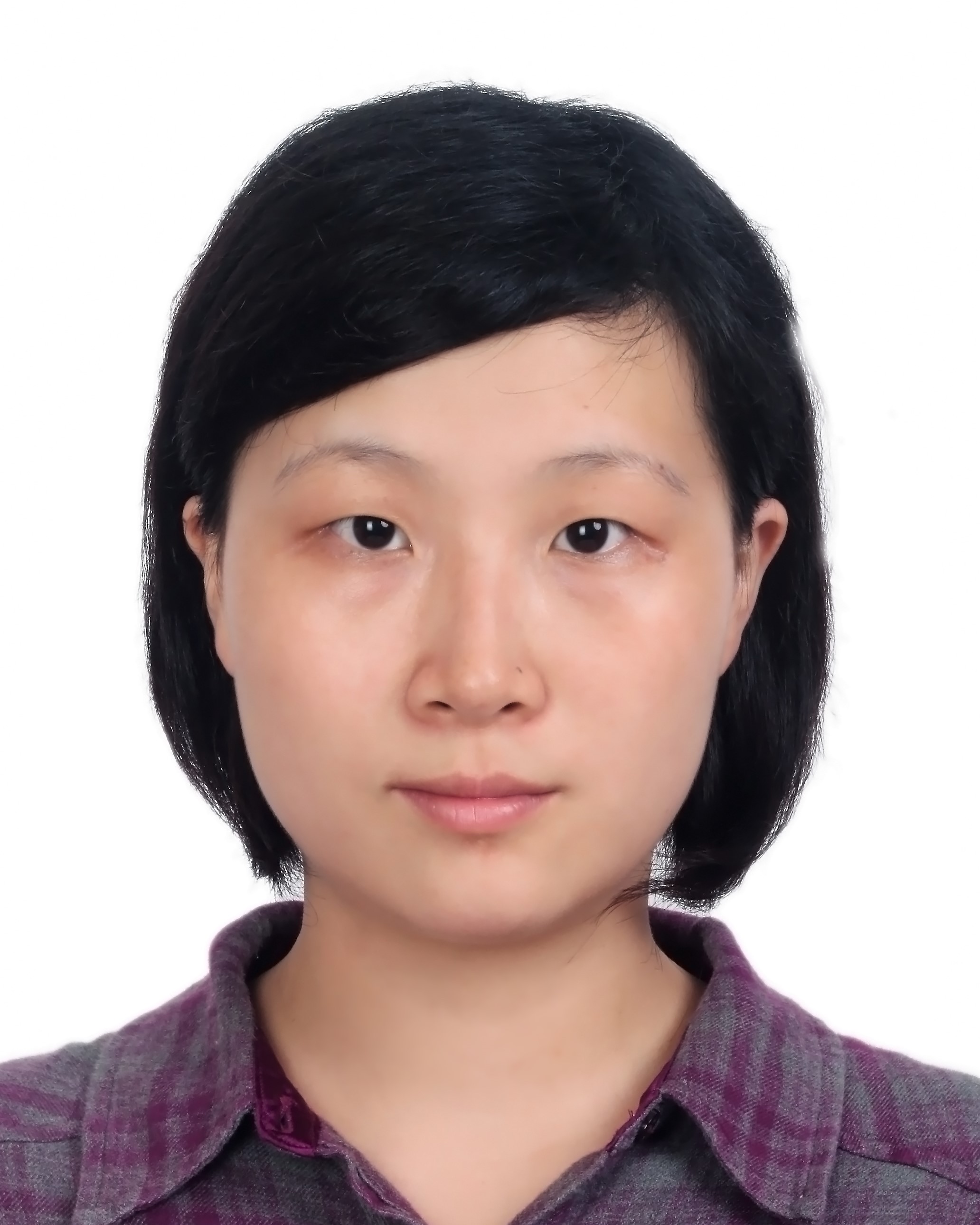}}]{Han Zhou} received the Ph.D. degree in control science and engineering from the National University of Defense Technology (NUDT), Changsha, China, in 2015. 
		She is currently an associate professor with the College of Intelligence Science and Technology, National University of Defense Technology. Her research interests include biomimetic robotics, collective intelligence, and learning control.
	\end{IEEEbiography}
	
	\begin{IEEEbiography}[{\includegraphics[width=1in,height=1.25in, clip,keepaspectratio]{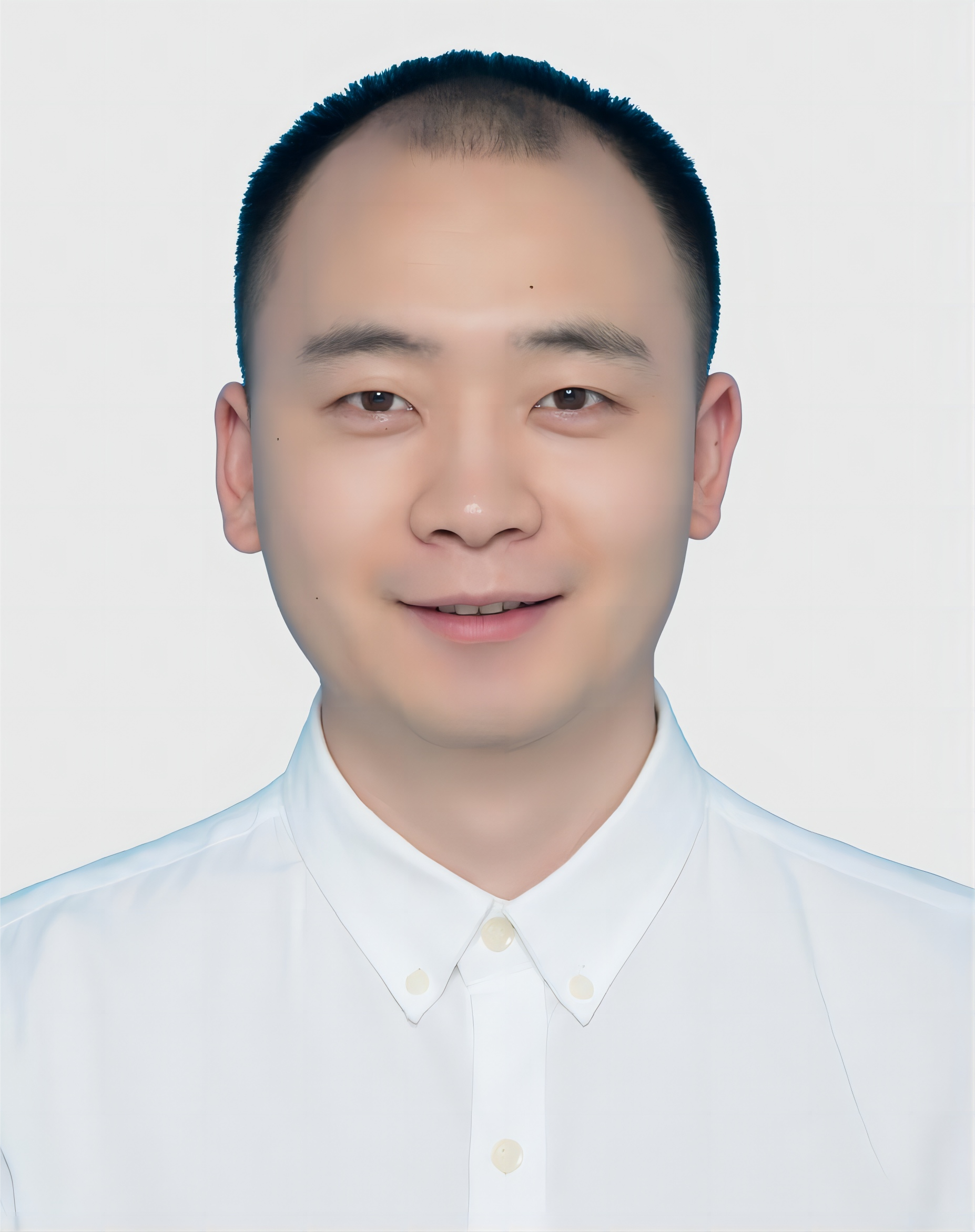}}]{Jun Lai} received the B.E., and Ph.D. degrees in instrument science and technology from the National University of Defense Technology (NUDT), Changsha, China, in 2013, and 2019, respectively.
		He is currently an assistant professor  with the College of Intelligence Science and Technology, NUDT. He is currently in the field of cooperative localization and cooperative mission planning of unmanned system.
	\end{IEEEbiography}
	
	\begin{IEEEbiography}[{\includegraphics[width=1in,height=1.25in, clip,keepaspectratio]{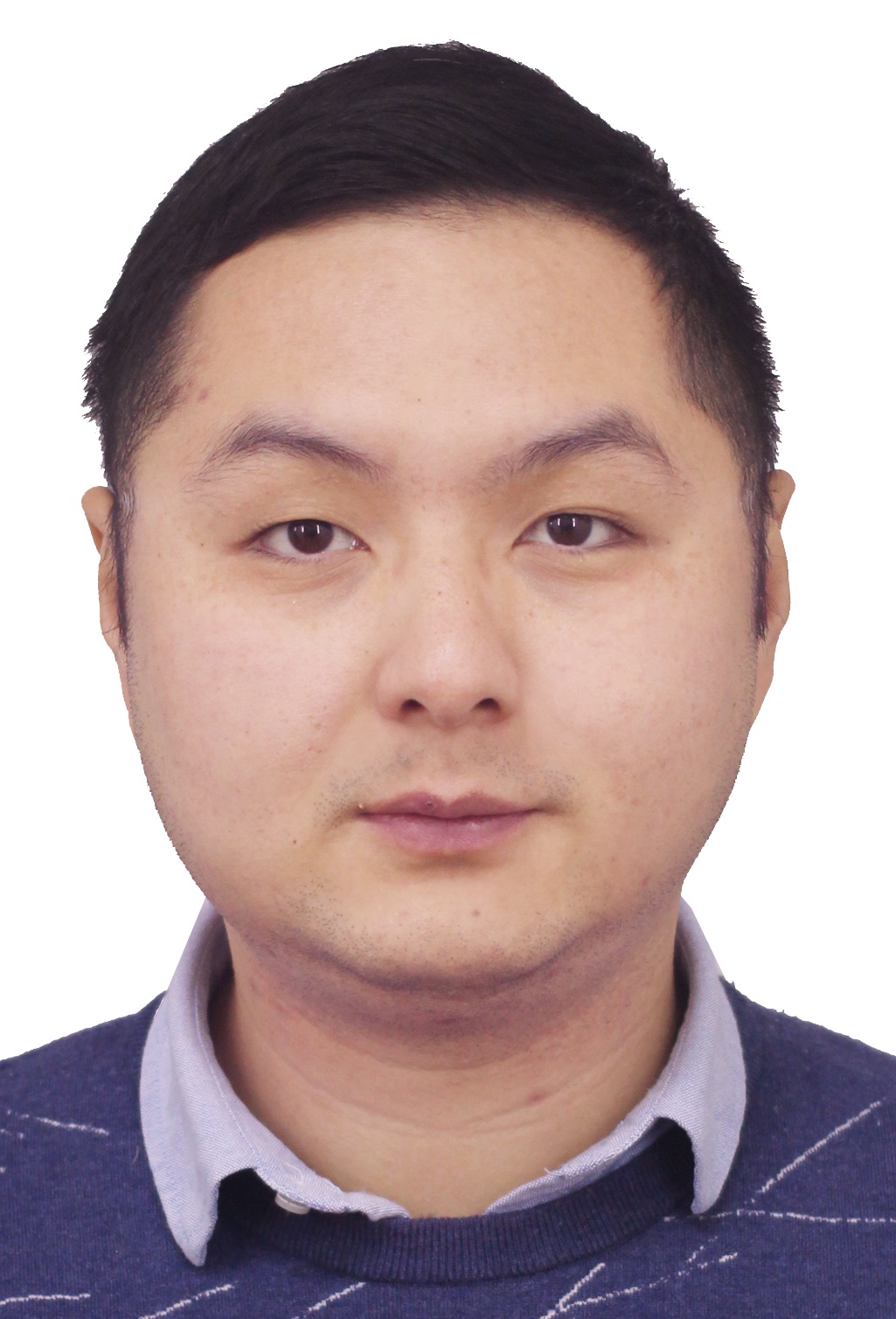}}]{Dengqing Tang} received the B.Eng., M.S. and Ph.D. degrees in control science and engineering from National University of Defense Technology, Changsha, China, in 2013, 2016 and 2019, respectively. 
		He is currently an Associate Professor in the College of Intelligence Science and Engineering, National University of Defense Technology. His research interests cover visual object detection, visual object pose estimation and deep learning.
	\end{IEEEbiography}
	
\end{document}